\documentclass[a4paper,fleqn,final]{cas-dc}

\usepackage[numbers]{natbib}
\usepackage{color}
\usepackage{xcolor}
\usepackage{amssymb}
\usepackage{supertabular}
\usepackage{multicol}
\usepackage{multirow}
\usepackage{subcaption}
\usepackage{adjustbox}
\usepackage{makecell}
\usepackage{amsmath}
\usepackage[figuresright]{rotating}
\usepackage{amssymb}
\usepackage{multirow}
\usepackage{booktabs}
\usepackage{hyperref}
\usepackage{pifont}
\hypersetup{
    colorlinks=true,
    linkcolor=red,
    citecolor=cyan,
    filecolor=magenta,      
    urlcolor=magenta,
}
\usepackage[table,dvipsnames]{xcolor}
\usepackage{float}
\usepackage{siunitx}

\usepackage[table]{xcolor}
\usepackage{colortbl}
\usepackage{multirow}
\usepackage{makecell}

\usepackage{bm}
\usepackage{pifont}
\def\tsc#1{\csdef{#1}{\textsc{\lowercase{#1}}\xspace}}
\tsc{WGM}
\tsc{QE}

\makeatletter
\def\ps@first{%
  \let\@oddhead\@empty
  \let\@evenhead\@empty
  \let\@oddfoot\@empty
  \let\@evenfoot\@empty
}
\def\ps@cas{%
  \let\@oddhead\@empty
  \let\@evenhead\@empty
  \let\@oddfoot\@empty
  \let\@evenfoot\@empty
}
\makeatother

\begin{document}
\pagestyle{cas}

\let\WriteBookmarks\relax
\def\floatpagepagefraction{1}
\def\textpagefraction{.001}

\shorttitle{}    

\shortauthors{}  

\title [mode = title]{M$^3$Net: A Macro$\rightarrow$Meso$\rightarrow$Micro Clinical-inspired hierarchical 
3D Network for Pulmonary Nodule Classification}  



%
\author[1,2]{Jinyue Li}
[orcid=0009-0003-2565-3243]


\credit{Data curation, Methodology, Visualization, Writing - original draft}

\affiliation[1]{organization={Hefei Cancer Hospital of CAS, Institute of Health and Medical Technology, Hefei Institutes of Physical Science, Chinese Academy of Sciences},
            city={Hefei},
            postcode={230031},
            country={PR China}}

\affiliation[2]{organization={University of Science and Technology of China},
            city={Hefei},
            postcode={230026},
            country={PR China}}

\author[1,2]{Yuzhou Yu}
[orcid=0009-0005-1844-8625]


\credit{Validation, Visualization}

\author[3,4]{Jingjing Yang}
[orcid=0009-0007-5246-7322]


\credit{Data curation, Formal analysis}

\affiliation[3]{organization={Graduate School, Bengbu Medical College},
            city={Bengbu},
            postcode={233030},
            country={PR China}}

\affiliation[4]{organization={Department of Pulmonary and Critical Care Medicine, The First Affiliated Hospital of USTC, Division of Life Sciences and Medicine, University of Science and Technology of China (USTC)},
            city={Hefei},
            postcode={230001},
            country={PR China}}

\author[4]{Meng Fu}
[orcid=0000-0002-7867-1168]


\credit{Conceptualization, Data curation, Formal analysis}

\author[1,2]{Yani Zhang}
[orcid=0000-0003-4545-5052]


\credit{Formal analysis}

\author[7]{Shuyao He}
[orcid=0009-0000-7870-9181]


\credit{Supervision, Validation, Writing - review \& editing}

\affiliation[7]{organization={Northeastern University},
            city={Boston},
            state={MA},
            postcode={02115},
            country={USA}}

\author[1]{Dianlong Ge}
[orcid=0000-0002-0706-4782]


\credit{Investigation, Methodology, Project administration}

\author[5]{Xin Ning}
[orcid=0000-0001-7897-1673]

\cormark[1]

\ead{ningxin@semi.ac.cn}

\credit{Investigation, Methodology, Project administration, Writing - review \& editing}

\affiliation[5]{organization={Institute of Semiconductors, Chinese Academy of Sciences},
            city={Beijing},
            postcode={100083},
            country={PR China}}

\author[1]{Yannan Chu}
[orcid=0000-0001-7422-905X]

\cormark[1]

\ead{ychu@aiofm.ac.cn}

\credit{Conceptualization, Data curation, Funding acquisition, Project administration}

\author[6]{Qiankun Li}
[orcid=0000-0001-5121-1682]


\credit{Resources, Software, Supervision, Validation, Visualization, Writing - original draft, Writing - review \& editing}

\affiliation[6]{organization={College of Computing and Data Science (CCDS), Nanyang Technological University},
            postcode={639798},
            country={Singapore}}

\cortext[1]{Corresponding authors.}



\begin{abstract}
The accurate classification of benign and malignant pulmonary nodules in CT scans is critical for early lung cancer screening, yet remains challenging due to the multi-scale and heterogeneous nature of pulmonary nodules.
While deep learning offers potential for auxiliary diagnosis, most existing models act as "black boxes", lacking the transparency and explainability required for trustworthy clinical integration. {
To address this issue, we propose M$^3$Net, a novel 3D network for pulmonary nodule classification inspired by the hierarchical diagnostic workflow of radiologists, which integrates multi-scale contextual information from fine-grained structures to global anatomical relationships.
}
Our framework constructs a progressive multi-scale input, from fine-grained nodule structures to local semantics and global spatial relationships. M$^3$Net employs scale-specific encoders and ensures cross-scale semantic consistency through latent space projection and mutual information maximization. Extensive experiments on the public LIDC-IDRI dataset and a self-collected clinical dataset (USTC-FHLN) demonstrate that our method achieves state-of-the-art performance, with accuracies of $\mathbf{86.96\%}$ and $\mathbf{84.24\%}$ respectively, outperforming the best baseline by $\mathbf{3.26\%}$ and $\mathbf{2.17\%}$. {The results validate that M$^3$Net provides a more robust and clinically relevant solution for pulmonary nodule classification.}
The code is available at \url{https://github.com/jylEcho/M3-Net}.
\end{abstract}


\begin{highlights}
\item A clinically guided Macro-Meso-Micro ($M^3$) 3D network enhances explainability in nodule classification.
\item Progressive multi-scale fusion aligns model reasoning with radiologists' hierarchical logic.
\item Hierarchical cross-attention enables progressive reasoning across local and global cues.
\item Experiments on LIDC-IDRI and clinical data show state-of-the-art performance and robustness.
\end{highlights}


\maketitle

\section{Introduction}
\label{sec:introduction}
In the clinical context of early lung cancer screening, the differentiation of benign and malignant pulmonary nodules represents a core challenge in radiologists' decision-making \cite{prosper2023expanding, xie2018fusing}. Chest computed tomography (CT), with its high spatial resolution and non-invasive nature, has become the cornerstone for early lung cancer screening and nodule assessment \cite{ohno2022efficacy}. However, the large-scale screening of benign and malignant nodules on CT images remains a complex and time-consuming task for radiologists \cite{zhou2022cascaded}. Diagnostic uncertainty directly translates into clinical risks, as misdiagnosis and missed diagnosis of nodules with different pathological natures can lead to adverse outcomes \cite{lv2024novel}. 
{Consequently, there is a pressing clinical need for auxiliary diagnostic systems that can transcend individual experiential differences and
provide objective decision support,
thereby enhancing the reliability and robustness of overall diagnostic results \cite{zhu2018deeplung, eman2025emeralds,alsedais2025artificial,elsedais2024artificial}.
}

In high-risk medical diagnostic fields that heavily rely on professional expertise, medical decision-making is a complex process involving attribution, accountability, and subsequent intervention \cite{tilala2024ethical,babushkina2023we,jiang2021learning,su2025streamline}. Therefore, the requirements for an auxiliary diagnostic system extend beyond high statistical accuracy to include transparency and explainability of the decision-making process \cite{dumaev2024concept,choi2021reproducible,wang2024explainable,jiang2021learning}. Explainability refers to the ability to understand the rationale and key evidence behind an artificial intelligence model's decisions. 
{An interpretable pulmonary nodule classification system can 
relate model predictions to imaging characteristics:
from subtle lobulation and spiculation of the nodule, to its relationship with surrounding vessels, and further to its spatial mass effect within the entire lung lobe \cite{jiang2021learning, zhu2023explainable,wang2024explainable,wang2024towards,shen2019interpretable}.
}

{This ability to 
relate internal feature representations to clinically recognizable imaging features
provides physicians with a critical evaluation interface,
}
enabling them to {
understand the model's prediction process
} from a pathophysiological perspective, thereby facilitating effective human-machine cross-verification. 
{
Thus, explainability can serve as a bridge connecting data-driven models with clinical diagnosis and may support the development of more trustworthy AI-assisted diagnostic tools.
}

Although medical auxiliary diagnostic systems are rapidly evolving, with numerous deep learning-based studies demonstrating excellent performance on benchmark datasets such as LIDC-IDRI and LUNA16 \cite{fu2022semi, saied2023efficient} including 3D convolutional neural networks \cite{zhu2018deeplung}, attention-based feature aggregation \cite{zhao2023attentive, zhou2023medical}, graph neural networks, and multi-scale feature fusion \cite{lv2024novel, jiang2021learning} significant limitations remain in terms of trustworthiness and clinical applicability. Some methods have incorporated explainability and visualization mechanisms to enhance clinical trust, such as explainable frameworks and attention heatmaps \cite{zhu2023explainable, he2022ishap}. Other studies have attempted to integrate multi-modal data, combining radiomics, molecular, or clinical features to improve classification robustness \cite{wang2022integrative, he2023accurate}. However, these approaches still face fundamental constraints. Firstly, most models rely solely on single-modal CT images, failing to capture the complete multi-dimensional characteristics of nodules, which leads to unstable performance in complex scenarios like ground-glass nodules \cite{zhang2024deep}. Secondly, these models often operate as "black boxes." Even with the introduction of explainability techniques like attention mechanisms, they fail to systematically model nodule semantics, boundary evolution, and multi-scale contextual cues from the surrounding lung tissue; their decision logic remains disconnected from the cognitive workflow of physicians \cite{wang2024explainable, zhang2024deep}. These shortcomings collectively hinder the system's ability to provide convincing decision support in real-world clinical diagnostics. Therefore, developing an auxiliary diagnostic framework that possesses both powerful representational capacity and the ability to provide hierarchical, clinically logical explanations has become the key to bridging the gap between AI diagnosis and clinical practice.

{Motivated by the hierarchical diagnostic workflow commonly adopted by radiologists, we first systematically evaluated the impact of input scales on the performance of benign and malignant pulmonary nodule classification. This analysis establishes a progressive hierarchical structure: \textbf{from fine-grained structure $\to$ local context $\to$ global spatial relationships}. Based on this observation, we propose M$^3$Net, a macro–meso–micro hierarchical 3D network for pulmonary nodule classification. By constructing progressive multi-scale inputs ranging from local fine structures to global spatial relationships, M$^3$Net integrates complementary information across different spatial contexts. Specifically, M$^3$Net employs scale-specific encoders to extract feature representations at each scale, and encourages cross-scale consistency through latent space projection, InfoNCE-based mutual information maximization, and second-order covariance alignment. Through hierarchical interaction and feature fusion, the model progressively aggregates information from local textures, surrounding tissues, and global anatomical context. 

Extensive experiments conducted on the public LIDC-IDRI dataset and a clinically collected dataset (USTC-FHLN) demonstrate that our method achieves state-of-the-art classification performance. The proposed approach achieves accuracies of $\mathbf{86.96\%}$ and $\mathbf{84.24\%}$ respectively, outperforming the best baseline by $\mathbf{3.26\%}$ and $\mathbf{2.17\%}$, indicating strong performance and generalization ability.}

The main contributions are summarized as follows:
\begin{itemize}
    \item We analyze the effect of input scale on lung nodule classification, showing that small-, medium-, and large-scale inputs respectively capture fine details, local context, and long-range semantic cues.
    \item {We propose M$^3$Net, a macro–meso–micro hierarchical 3D network that integrates multi-scale information ranging from local fine structures to global spatial context.}  
    \item Extensive experiments on two benchmark datasets (LIDC-IDRI and USTC-FHLN) show that the proposed method achieves state-of-the-art classification performance.	
\end{itemize}

The remainder of this paper is structured as follows. Section \ref{section:related work} reviews related work, while Section \ref{section:method} details the architecture and methodological framework of the proposed M$^3$Net. Section \ref{sec:Experiments} describes the experimental setup, including datasets, evaluation criteria, and implementation settings. Section \ref{section:Experimental Results} presents the results and analysis, demonstrating the performance and robustness of the proposed method. Finally, Section \ref{section:conclusion} concludes the paper and discusses future work, followed by acknowledgments in Section \ref{sec:acknowledge}.

\section{Related Work}\label{section:related work}

{
\subsection{Methodological Advances in Nodule Classification}
Recent years have witnessed remarkable methodological progress in pulmonary nodule classification driven by deep learning \cite{wani2024explainable,xie2018knowledge,mobiny2021memory}.
While early studies primarily focused on improving predictive accuracy, recent research has increasingly emphasized the interpretability and clinical transparency of these models \cite{xie2019semi,yan2024tdf}.
Consequently, the design of architectures capable of learning discriminative yet interpretable representations from limited and heterogeneous CT data has become a central research focus\cite{longo2024explainable,brankovic2025clinician,jin2023guidelines}.
}

\subsubsection{Attention enhanced and explainability-driven methods}
To enhance representation diversity, multi-scale and multi-view learning frameworks have been widely explored. Lv et al.~\cite{lv2024novel} proposed a lightweight multi-scale interleaved fusion network (MIFNet) to efficiently capture hierarchical spatial semantics while maintaining only 0.7M parameters. Zhou et al.~\cite{zhou2022cascaded} developed a 2.5D cascaded multi-stage framework for nodule detection and segmentation, improving sensitivity and computational efficiency. Similarly, Xiong et al.~\cite{xiong2024pulmonary} adopted a model-fusion-based 2.5D architecture combining weight box fusion and adaptive 3D CNNs to achieve a 97.4\% sensitivity rate on LUNA16, and Tang et al.~\cite{tang2019nodulenet} integrated detection, false-positive reduction, and segmentation into a unified multitask 3D network (NoduleNet). These approaches collectively demonstrate the effectiveness of fusing multi-scale contextual information for accurate nodule assessment.

{
\subsubsection{Explainability enhanced global–local attention methods}
Longo et al.~\cite{longo2024explainable} presented a comprehensive manifesto outlining 28 open problems across nine categories to advance Explainable AI (XAI) through interdisciplinary collaboration and real-world alignment. Brankovic et al.~\cite{brankovic2025clinician} developed the clinician-informed CLIX-M, a 14-item checklist with metrics to standardize the evaluation and reporting of XAI in clinical decision support systems. Jin et al.~\cite{jin2023guidelines} proposed Clinical XAI Guidelines with five optimization criteria and demonstrated through systematic evaluation that many existing heatmap-based methods fail to meet clinical requirements, highlighting the need for clinically grounded assessment frameworks.
Yi et al.~\cite{yi2022multi} proposed a multi-label softmax loss network (MLSL-Net) to model label dependencies, and He et al.~\cite{he2022ishap} utilized an ISHAP-based interpretable classifier that leverages semantic and radiomics features to improve transparency and reliability. Tang et al.~\cite{tang2020automated} also validated the potential of CNN-based feature extraction in chest radiography for general disease classification, providing methodological insights for pulmonary tasks.

}

\subsubsection{Semi-supervised and generative methods}
Roy et al.~\cite{roy2022adgan} proposed an attribute-driven generative adversarial network (ADGAN) incorporating self-attention U-Net modules for multiclass classification, while Fu et al.~\cite{fu2022semi} introduced a reverse adversarial classification network (RACN) to diagnose five pathological subtypes using semi-supervised learning. These methods effectively leverage unlabeled or synthetic data to improve generalization. Furthermore, Zhu et al.~\cite{zhu2018deeplung} developed DeepLung, a fully automated 3D DPN-based system combining detection and classification in an end-to-end manner, achieving expert-level accuracy.

\subsection{Clinical-Oriented Advances in Nodule Classification}

Recent research increasingly emphasizes comprehensive clinical intelligence \cite{wang2024multi}, seeking to bridge imaging, molecular, and clinical domains for more accurate malignancy characterization and real-world relevance \cite{gupta2024texture,gunawan2024combining}.

\subsubsection{Multi-modal learning methods}
Wang et al.~\cite{wang2022integrative} proposed a metabolic–protein neural network (MP-NN) that integrates serum metabolic fingerprints with tumor marker CEA, while the extended MPI-RF model further incorporated image features for early lung adenocarcinoma detection. Liu et al.~\cite{liu2024lung} demonstrated that incorporating fibrotic microenvironment and semantic fibrosis metadata into CNNs significantly enhances malignancy classification performance. Tong et al.~\cite{tong2020pulmonary} fused CT-derived 3D-ResNet features with clinical data via multi-kernel learning, providing a hybrid radiology–clinical framework for improved decision-making.


\subsubsection{Efficiency and explainability in diagnostic pipelines}
From a population and systems perspective, Prosper et al.~\cite{prosper2023expanding} reviewed methodological challenges in translating deep learning into clinical workflows, emphasizing issues of heterogeneity, annotation quality, and standardization. Zhang et al.~\cite{zhang2024deep} trained four deep learning models for malignancy risk estimation of sub-centimeter nodules, achieving AUC values up to 0.942 and performance comparable to senior clinicians. These results suggest that AI systems can augment human expertise in risk stratification and follow-up management.
Additionally, Tang et al.~\cite{tang2020automated} and Zhou et al.~\cite{zhou2023medical} emphasized generalizability and clinical efficiency in imaging diagnosis, while Shen et al.~\cite{shen2017multi} pioneered multi-crop CNNs capable of modeling malignancy suspicion directly from raw CT patches without segmentation. Collectively, these studies signify a paradigm shift toward clinically guided, data-driven, and interpretable frameworks that integrate multi-source medical information for precise pulmonary nodule management.

{
In summary, research related to model interpretation in pulmonary nodule classification has primarily followed two paradigms: one improves transparency through architectural innovations like multi-scale fusion \cite{lv2024novel, zhou2022cascaded} and built-in attention mechanisms \cite{zhou2023medical}, implicitly guiding the model to focus on salient regions; the other relies on post-hoc techniques, such as multi-modal data integration \cite{wang2022integrative} or feature visualization \cite{zhu2023explainable}, to justify decisions after they are made. While valuable, these approaches offer explainability that is either indirect or functionally separated from the model’s feature representations.

Despite these advances, interpreting deep learning models for pulmonary nodule classification remains challenging. In many existing approaches, explainability remains an external, post-hoc attribute rather than an inherent property of the model’s internal feature representations. 
These methods either perform post-training attribution analysis on “black-box” models or produce isolated saliency maps, which may not fully reflect how information is integrated across different spatial contexts.
This results in fragmented explanations detached from the decision process, making it difficult for physicians to interpret model predictions from a clinical perspective.
}

{Different from these existing methods, this paper proposes 
M$^3$Net: A Macro$\rightarrow$Meso$\rightarrow$Micro Clinical-inspired hierarchical 3D Network
The proposed framework integrates multi-scale contextual information within the model architecture, enabling progressive feature aggregation from fine-grained structures to broader anatomical context.
}

\begin{figure*}[h]
\centering
\includegraphics[width=\textwidth]{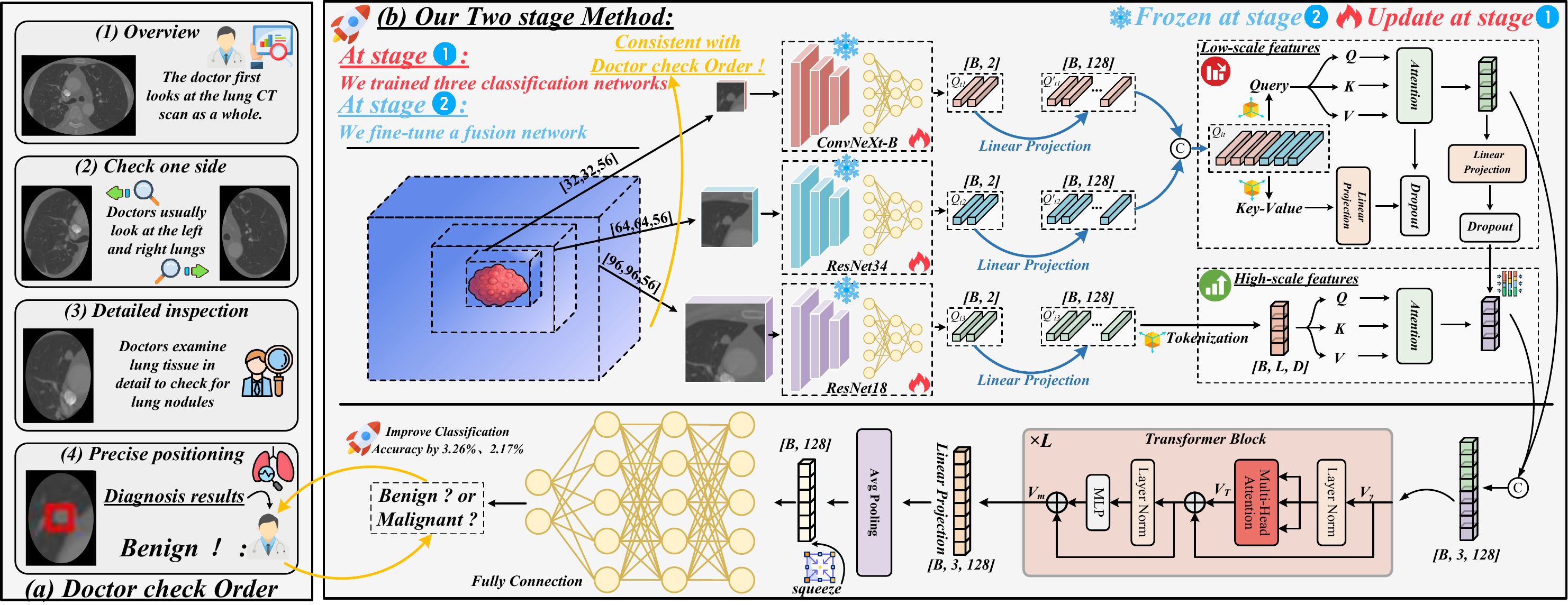}
\caption{{Overview of M$^3$Net. (a) Radiologists’ diagnostic workflow motivating the two-stage framework.
(b) The proposed method exploits neighborhood information of lung nodules through three independently fine-tuned classifiers with complementary context.}}
\label{Figure1}
\end{figure*}

\section{Methodology}\label{section:method}

\subsection{Overall Framework $\&$ Clinical Interpretation and Value}

In clinical radiology, pulmonary nodules are evaluated through a hierarchical cognitive process (Figure~\ref{Figure1} (a)): (1) \textbf{\textit{global screening}} to localize suspicious regions, (2) \textbf{\textit{boundary inspection}} to evaluate margin sharpness and lobulation, and (3) \textbf{\textit{fine-grained internal assessment}} for malignancy confirmation. 
This diagnostic reasoning integrates multi-scale visual perception and context-dependent decision refinement.

Inspired by this process, each fusion level of our framework corresponds to a distinct layer of clinical cognition:  
\begin{itemize}
    \item \textbf{Macro-scale} represents \textbf{\textit{holistic synthesis}}, akin to the clinician’s comprehensive judgment integrating multi-level cues.
    \item \textbf{Meso-scale} mirrors \textbf{\textit{contextual integration}}, linking nodular appearance with surrounding parenchymal or vascular structures.
    \item \textbf{Micro-scale} emulates \textbf{\textit{edge reasoning}}, capturing spiculation and lobulation at the nodule margin.
\end{itemize}

Consequently, our model \textbf{translates radiological heuristics into a mathematically structured multi-scale learning paradigm.} 
By aligning its computational inference with human diagnostic logic, 
it enhances both classification accuracy and clinical explainability, supporting trustworthy computer-aided diagnosis.

Formally, we represent each 3D nodule using three nested volumetric patches:
\begin{equation}
\mathcal{X} = \{\mathbf{X}_{96},\, \mathbf{X}_{64},\, \mathbf{X}_{32}\}, \quad 
\mathbf{X}_{s} \in \mathbb{R}^{s \times s \times 56},
\end{equation}
representing \textbf{\textit{global}}, \textbf{\textit{peri-nodular}}, and \textbf{\textit{local contexts}}, respectively. 
Such scale decomposition embeds the radiologist’s hierarchical reasoning into the learning process, enabling the model to jointly capture intrinsic morphology and extrinsic anatomical priors that are critical for malignancy discrimination.

\subsection{Scale-Specific Encoding and Representation Alignment}

Each cropped volume $\mathbf{X}_s$ is independently processed by a scale-specific backbone network $\mathcal{B}_s$, which learns discriminative representations tailored to its spatial scale:
\begin{equation}
\mathbf{F}_s = \mathcal{B}_s(\mathbf{X}_s; \theta_s), 
\quad 
\mathbf{F}_s \in \mathbb{R}^{B \times C_s},
\end{equation}
where $B$ denotes the batch size and $C_s$ the channel dimension at scale $s$.  
{
To encourage both scale-specific expressivity and scale-invariant stability, 
we adopt a two-stage training protocol.
In the first stage, each backbone is pre-trained under a supervised 
classification objective:
\begin{equation}
\min_{\theta_s,\phi_s}\ 
\mathcal{L}_{\mathrm{cls}}^{(s)}
=
\mathbb{E}_{(\mathbf{X}_s,y)}
\left[
\mathcal{L}_{\mathrm{CE}}
\big(
y,\,
\Phi_s(\mathbf{F}_s)
\big)
\right],
\end{equation}
where $\mathbf{F}_s = f_{\theta_s}(\mathbf{X}_s)$ denotes the feature 
representation extracted by the $s$-th backbone, 
$\Phi_s(\cdot)$ is a lightweight classification head parameterized by 
$\phi_s$, and $\mathcal{L}_{\mathrm{CE}}$ denotes the standard cross-entropy loss.
After convergence, we partially freeze the lower convolutional layers 
in $\theta_s$ to preserve low-level texture and edge representations, 
while keeping higher-level layers trainable for subsequent semantic adaptation.
}
\paragraph{\textbf{Latent projection}}  
All scale-specific features are projected into a unified latent space $\mathcal{Z}\subset\mathbb{R}^D$ to enable semantic alignment across scales:
\begin{equation}
\mathbf{Z}_s = \mathbf{F}_s\mathbf{W}_s + \mathbf{b}_s,
\quad 
\mathbf{Z}_s \in \mathbb{R}^{B\times D},
\end{equation}
where $\mathbf{W}_s\!\in\!\mathbb{R}^{C_s\times D}$ and $\mathbf{b}_s\!\in\!\mathbb{R}^D$.  
This shared embedding space facilitates scale-invariant reasoning and enables consistent feature comparison across different anatomical contexts.

\paragraph{\textbf{Mutual-information maximization}}  
To enforce semantic consistency between different scales, we maximize the mutual information between latent embeddings.  
Since direct estimation of $I(\mathbf{Z}_i;\mathbf{Z}_j)$ is intractable, we adopt the InfoNCE contrastive formulation as a variational lower bound:
\begin{equation}
\mathcal{L}_{\text{InfoNCE}}^{(i,j)}
=-\mathbb{E}\!
\left[
\log
\frac{\exp\!\big(\mathrm{sim}(\mathbf{z}_i,\mathbf{z}_j^+)/\tau\big)}
{\sum_{k=1}^{N}\exp\!\big(\mathrm{sim}(\mathbf{z}_i,\mathbf{z}_k)/\tau\big)}
\right],
\end{equation}
where $\mathrm{sim}(\mathbf{u},\mathbf{v})=\mathbf{u}^\top\mathbf{v}/(\|\mathbf{u}\|\|\mathbf{v}\|)$ denotes cosine similarity, $\tau$ is the temperature parameter, and $N$ is the number of negative pairs in the batch.  
By minimizing $\mathcal{L}_{\text{InfoNCE}}$, we maximize a lower bound of the mutual information between $\mathbf{Z}_i$ and $\mathbf{Z}_j$.

\paragraph{\textbf{Second-order alignment and subspace regularization}}  
Beyond contrastive consistency, we additionally constrain the second-order feature statistics to align across scales.  
Let $\Sigma_s = \mathrm{Cov}(\mathbf{Z}_s)$ denote the covariance matrix of scale $s$.  
We minimize the Frobenius distance between scale-wise covariances:
\begin{equation}
\mathcal{L}_{\text{cov}} 
= \sum_{i<j} \|\Sigma_i - \Sigma_j\|_F^2,
\end{equation}
which stabilizes feature geometry and suppresses scale-induced deformation.  
To prevent degenerate solutions (e.g., 
correlated embeddings), we introduce orthogonality and low-rank priors:
\begin{equation}
\scalebox{1.0}{$
\begin{gathered}
\mathcal{L}_{\text{orth}} = \sum_s \|\mathbf{Z}_s^\top\mathbf{Z}_s - \mathbf{I}\|_F^2,
\quad
\mathcal{L}_{\text{nuc}} = \sum_s \|\mathbf{Z}_s\|_*,
\end{gathered}
$}
\end{equation}
where $\|\cdot\|_*$ is the nuclear norm that promotes compact and complementary subspaces across scales.
{
\paragraph{\textbf{Overall alignment objective}}

To encourage consistent representations across scales, we introduce a hierarchical alignment loss defined as

\begin{equation}
\scalebox{1.0}{$
\begin{gathered}
\mathcal{L}_{\text{align}} =
\sum_{i<j}\lambda_{\text{NCE}}^{(i,j)}\mathcal{L}_{\text{InfoNCE}}^{(i,j)}
+ \lambda_{\text{cov}}\mathcal{L}_{\text{cov}}
+ \lambda_{\text{orth}}\mathcal{L}_{\text{orth}}
+ \lambda_{\text{nuc}}\mathcal{L}_{\text{nuc}},
\end{gathered}
$}
\end{equation}
where the InfoNCE term promotes cross-scale feature consistency, $\mathcal{L}_{\mathrm{align}}$ is the cross-scale representation alignment loss, while the covariance, orthogonality, and nuclear norm regularizers encourage feature diversity and prevent representation collapse.

The model is trained in two stages.

\textbf{Stage 1 (Discriminative pretraining).}  
Each scale-specific branch is first optimized independently using the classification objective

\begin{equation}
\min_{\{\theta_s,\mathbf{W}_s,\mathbf{b}_s\}}
\sum_s \mathcal{L}_{\mathrm{CE}}^{(s)},
\end{equation}
where $\mathcal{L}_{\mathrm{CE}}^{(s)}$ denotes the cross-entropy classification loss for the $s$-th scale branch. This stage learns stable discriminative representations for each scale-specific branch.

\textbf{Stage 2 (Cross-scale alignment).}  
Starting from the pretrained model, we further optimize the network using the joint objective

\begin{equation}
\min_{\{\theta_s,\mathbf{W}_s,\mathbf{b}_s\}}\;
\sum_s \mathcal{L}_{\mathrm{CE}}^{(s)}
+ \beta\,\mathcal{L}_{\mathrm{align}},
\end{equation}
where $\beta$ is a trade-off coefficient that balances the classification objective and the alignment objective. Specifically, $\mathcal{L}_{\mathrm{align}}$ consists of the InfoNCE-based cross-scale consistency term and three regularization terms, weighted by $\lambda_{\text{NCE}}^{(i,j)}$, $\lambda_{\text{cov}}$, $\lambda_{\text{orth}}$, and $\lambda_{\text{nuc}}$, respectively.

This hierarchical training strategy preserves stable low-level texture and boundary cues learned during Stage~1, while progressively enforcing semantic consistency across scales during Stage~2.
}
\subsection{Hierarchical Cross-Attention Fusion}

After alignment, we fuse the multi-scale embeddings via a hierarchical cross-attention mechanism emulating radiologists’ progressive diagnostic reasoning, from local texture inspection to contextual understanding and final holistic decision.

\paragraph{\textbf{Multi-head cross-attention}}  
Given query key value triplets $(\mathbf{Q},\mathbf{K},\mathbf{V})$ obtained through learned projections:
\begin{equation}
\mathbf{Q} = \mathbf{Z}\mathbf{W}_Q,\quad
\mathbf{K} = \mathbf{Z}\mathbf{W}_K,\quad
\mathbf{V} = \mathbf{Z}\mathbf{W}_V,
\end{equation}
the multi-head attention operation is formulated as:
\begin{equation}
\mathrm{Attn}(\mathbf{Q},\mathbf{K},\mathbf{V})
=\mathrm{softmax}\!\left(
\frac{\mathbf{Q}\mathbf{K}^\top}{\sqrt{d_h}}
\right)\mathbf{V},
\end{equation}
where $d_h$ denotes the per-head dimensionality.  
To enhance discriminability, a learnable temperature $\tau_h$ can be incorporated to adaptively scale attention sharpness:
\begin{equation}
\mathrm{Attn}_h(\mathbf{Q},\mathbf{K},\mathbf{V})
=\mathrm{softmax}\!\left(
\frac{\mathbf{Q}\mathbf{K}^\top}{\tau_h\sqrt{d_h}}
\right)\mathbf{V}.
\end{equation}

\paragraph{\textbf{Stage I: Peripheral reasoning}}  
Low- and medium-scale embeddings exchange boundary and texture cues through the first cross-attention layer:
\begin{equation}
\label{eq:transformer_block1}
\mathbf{T}_1
=\mathrm{CrossAttn}\!\left(
\mathbf{Z}_{32}\mathbf{W}_Q^{(1)},
\mathbf{Z}_{64}\mathbf{W}_{KV}^{(1)}
\right),
\end{equation}
which mimics how radiologists correlate local morphology with adjacent tissues to assess structural integrity.

{
\paragraph{\textbf{Stage II: Context assimilation}}  
Let $B$ denote the batch size, $L$ the number of spatial tokens (i.e., flattened patch embeddings), and $D$ the embedding dimension. The global-scale embedding $\mathbf{Z}_{96}\!\in\!\mathbb{R}^{B\times L\times D}$ interacts with the refined features $\mathbf{T}_1$ through a second cross-attention module:
\begin{equation}
\label{eq:transformer_block2}
\mathbf{T}_2
=\mathrm{CrossAttn}\!\left(
\mathbf{Z}_{96}\mathbf{W}_Q^{(2)},
\mathbf{T}_1\mathbf{W}_{KV}^{(2)}
\right),
\end{equation}
where $L$ denotes the token sequence length obtained by flattening the spatial feature map into patch embeddings.
This design allows large-scale contextual knowledge to modulate fine-grained local cues through token-level interaction.
}

\paragraph{\textbf{Stage III: Hierarchical Transformer integration}}  
The concatenated sequence $\mathbf{T}=[\mathbf{T}_1;\mathbf{T}_2]$ is passed through a hierarchical Transformer encoder with residual refinement:
\begin{equation}
\mathbf{H}
=\mathrm{LN}\!\left(
\mathbf{T}
+ \mathcal{MHA}(\mathbf{T})
+ \mathcal{FFN}(\mathbf{T})
\right),
\end{equation}
where $\mathcal{MHA}$ and $\mathcal{FFN}$ denote multi-head attention and feed-forward submodules.  
We further impose an $\ell_2$ regularization term on attention weights to suppress over-confident focus:
\begin{equation}
\mathcal{L}_{\text{attn-reg}} = \sum_\ell \|\mathrm{Attn}^{(\ell)}\|_2^2.
\end{equation}

Finally, the fused features are globally aggregated and fed into a classifier to produce the final prediction:
\begin{equation}
\scalebox{1.0}{$
\begin{gathered}
\mathbf{h} = \mathrm{LN}\!\left(\mathrm{Pool}(\mathbf{H})\right), 
\\
\hat{y} = \mathrm{softmax}\!\left(
\mathbf{W}_c\,\mathrm{GELU}(\mathbf{W}_h\mathbf{h})
\right),
\end{gathered}
$}
\end{equation}
yielding the malignancy probability $\hat{y}\in[0,1]^2$.  
This hierarchical attention mechanism achieves progressive reasoning, first aligning fine-scale textures, then assimilating global semantics, and finally integrating both within a unified diagnostic representation.


\section{Experiments}\label{sec:Experiments}
\subsection{Dataset}
This study utilized two datasets: the publicly available Lung Image Database Consortium and Image Database Resource Initiative (LIDC–IDRI) dataset~\cite{armato2011lung}, and a collected clinical lung nodule CT (USTC-FHLN) dataset.
The detailed information of the two datasets is listed in Table~\ref{Dataset-details}.

\textbf{\ding{182} LIDC–IDRI Dataset.} 
The dataset collected retrospectively from seven academic institutions participating in the Lung Image Database Consortium and Image Database Resource Initiative (LIDC--IDRI). The scans were acquired using multiple CT scanner models from GE, Siemens, Philips, and Toshiba, under varying protocols, all data were de-identified according to HIPAA regulations and stored in DICOM format.
Each CT scan contained detailed lesion annotations generated through a two-phase, four-radiologist review process. In the first (blinded) phase, each radiologist independently marked lesions of interest, including nodules $\geq \SI{3}{mm}$, nodules $< \SI{3}{mm}$, and non-nodule findings $\geq \SI{3}{mm}$, without knowledge of other readers’ results. In the second (unblinded) phase, all radiologists reviewed each other’s anonymized annotations and optionally revised their own markings, producing a consensus-like multi-rater annotation scheme while preserving inter-reader variability.
Across all scans, 7,371 lesions were annotated as nodules by at least one radiologist, including 2,669 nodules $\geq \SI{3}{mm}$, of which 928 were consistently marked by all four readers. Each annotated nodule $\geq \SI{3}{mm}$ includes pixel-level contour delineations and a set of subjective feature ratings, including subtlety, shape, margin, spiculation, lobulation, internal texture, solidity, and likelihood of malignancy (scored 1--5). For 268 cases, retrospective pathologic diagnoses are available, indicating whether lesions correspond to nonmalignant, primary lung cancer, or metastatic disease.

{For label construction on LIDC-IDRI, malignancy scores (1–5) assigned independently by up to four radiologists were aggregated at the nodule level. The mean malignancy score across all available readers was computed, and binarized as follows: mean score $> 3$ was defined as malignant, while mean score $\leq$ 3 was defined as benign. No nodules were discarded due to indeterminate scores.}

\textbf{\ding{183} USTC-FHLN Dataset.} 
This study used a single-center, retrospective lung CT dataset collected from the First Affiliated Hospital of the University of Science and Technology of China. All data were de-identified in compliance with institutional requirements and stored in DICOM format.
{The dataset consists of CT scans from 97 patients, comprising a total of 7,967 axial images. The patient cohort included 68 males and 29 females, with ages ranging from 39 to 84 years (mean age 63.7 ± 10.1 years). Among them, 38 patients had a history of smoking, while 59 patients reported no smoking history. In addition, 23 patients had a history of alcohol consumption, whereas the remaining 74 patients reported no alcohol use.}
Scans were acquired using a NeuViz 128 CT scanner (Neusoft Medical Systems) under routine clinical protocols, with a slice thickness of 5.0 mm, tube voltage of 120 kV, tube current of 200 mA, and a lung reconstruction window (window width 1500 HU, window level -600 HU). The field of view was approximately 43.3 $\times$ 43.3 cm, reflecting standard thoracic imaging settings.
{A total of 224 pulmonary nodules were identified and annotated in the dataset, including 102 benign nodules and 122 malignant nodules. The diagnosis of malignant or benign nodules was primarily confirmed through histopathological examination. Pathological confirmation was obtained through bronchoscopic biopsy or CT-guided percutaneous lung biopsy, while a smaller subset of cases was confirmed through surgical resection and postoperative pathological analysis.}
{
Nodule annotations were performed at the pixel level using contour delineations. The labeling process relied on characteristic features observed in chest CT images to distinguish tumor regions from non-tumor regions. Initially, a radiologist manually delineated the nodule boundaries and assigned preliminary labels. Subsequently, the annotations were jointly reviewed and verified by one radiologist and one respiratory physician, ensuring the accuracy and clinical reliability of the segmentation masks and diagnostic labels.
}
This dataset provides expert-validated nodule-level segmentation masks and diagnostic labels, supporting the development and evaluation of automated lung nodule analysis methods in a real-world clinical setting.
{The USTC-FHLN dataset has received approval from the institutional ethics committee under certification number YD9110002062.}

\begin{figure*}[t]
\centering
\includegraphics[width=0.95\textwidth]{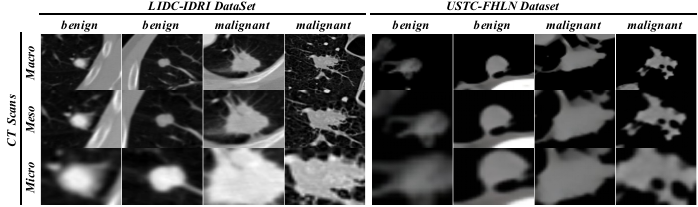}
\caption{Example images from the LIDC-IDRI and USTC-FHLN datasets.}
\label{Dataset_UNETR.png}
\end{figure*}

{
Following previous work \cite{wang2017chestx}, the public dataset, consisting of 2,625 pulmonary nodules (1,992 benign and 633 malignant), was randomly divided into 70\% for training, 10\% for validation, and 20
In addition, a private dataset comprising 224 nodules (102 benign and 122 malignant) was collected for external evaluation. To simulate the few-shot adaptation scenario, 20 positive–negative pairs (i.e., 40 nodules in total) were randomly selected for few-shot fine-tuning, while the remaining 184 nodules were used exclusively for inference and testing.
Figure \ref{Dataset_UNETR.png} shows example images from the datasets. The image dimensions and other relevant details are summarized in Table \ref{Dataset-details}.
}

\begin{table}[h]
\centering
\caption{Dataset characteristics}
{\renewcommand{\arraystretch}{1.15}
\resizebox{0.47\textwidth}{!}{
\begin{tabular}{ccc}
\Xhline{1.2pt}
\cellcolor{CadetBlue!20}\textbf{Dataset} & \cellcolor{CadetBlue!20}\textbf{Attribution} & \cellcolor{CadetBlue!20}\textbf{Value} \\
\Xhline{1.2pt}
\multirow{6}{*}{\shortstack{LIDC\\-IDRI}} 
 & CT numbers & 1,018 cases (from 1,010 subjects)  \\ 
\rowcolor{gray!10}
 & Slice thickness & \SIrange{}{3.0}{mm} (mean \SI{1.74}{mm}) \\
 & Slice resolution & \makecell{$512 \times 512$ matrix \\ 
\SIrange{0.46}{0.98}{mm} per pixel} \\
\rowcolor{gray!10}
 & Disease type & \makecell{Benign lesions, primary lung cancer, \\metastatic lesions} \\
\Xhline{1.2pt}
\multirow{6}{*}{\shortstack{USTC\\-FHLN}}  
 & CT numbers & 97 cases (from 97 subjects)  \\ 
\rowcolor{gray!10}
 & Slice thickness & 5.0 mm \\
 & Slice resolution & \makecell{$512 \times 512$ matrix \\ 0.85 mm per pixel} \\
\rowcolor{gray!10}
 & Disease type & \makecell{Benign lesions, Malignant lesions} \\
\Xhline{1.2pt}
\end{tabular}}
}
\label{Dataset-details}
\end{table}

\subsection{Evaluation Metrics}

To evaluate the performance of the lung nodule malignancy classification model, we employed several widely used classification metrics, including Accuracy (ACC), Balanced Accuracy (BACC), Precision (Pre), Recall (Rec), F1-score (F1), Specificity (Spec), Receiver Operating Characteristic Area Under Curve (ROC AUC), and Precision–Recall Area Under Curve (PR AUC). These metrics were derived from the confusion matrix consisting of true positives (TP), false positives (FP), true negatives (TN), and false negatives (FN).

Let $C$ denote the number of classes. For each class $c$, $TP_c$, $TN_c$, $FP_c$, and $FN_c$
denote the numbers of true positives, true negatives, false positives, and false negatives,
respectively. Let $N_c$ be the number of samples in class $c$, and the class weight is defined as
$ w_c = \displaystyle\frac{N_c}{\sum_{k=1}^{C} N_k} $.

The formulas for calculating weighted precision and recall are as follows:
$\mathrm{Pre} = \sum_{c=1}^{C} w_c \displaystyle\frac{TP_c}{TP_c + FP_c}$,
$\mathrm{Rec} = \sum_{c=1}^{C} w_c \displaystyle\frac{TP_c}{TP_c + FN_c}$,
specificity and F1 score are calculated as follows:
$\mathrm{Spec} = \sum_{c=1}^{C} w_c \displaystyle\frac{TN_c}{TN_c + FP_c}$, and
$\mathrm{F1} = \sum_{c=1}^{C} w_c
\displaystyle\frac{2 \cdot \mathrm{Pre}_c \cdot \mathrm{Rec}_c}
{\mathrm{Pre}_c + \mathrm{Rec}_c}$.
{We evaluate the classification performance using accuracy (ACC), balanced accuracy (BACC) \cite{nurzynska2023multilayer}, weighted ROC-AUC and PR-AUC:}

\begin{equation}
\mathrm{ACC} =
\frac{\sum_{c=1}^{C} (TP_c + TN_c)}
{\sum_{c=1}^{C} (TP_c + TN_c + FP_c + FN_c)},
\end{equation}

\begin{equation}
\mathrm{BACC} =
\frac{1}{C} w_c
\sum_{c=1}^{C}
\frac{TP_c}{TP_c + FN_c},
\end{equation}




\begin{equation}
\centering
\begin{aligned}
\mathrm{ROC\text{-}AUC} &=
\sum_{c=1}^{C} w_c
\int_{0}^{1} \mathrm{TPR}_c(\mathrm{FPR}_c)\, d(\mathrm{FPR}_c),
\end{aligned}
\end{equation} 

\begin{equation}
\centering
\begin{aligned}
\mathrm{PR\text{-}AUC} &=
\sum_{c=1}^{C} w_c
\int_{0}^{1}
\mathrm{Pre}(\mathrm{Rec})\, d(\mathrm{Rec}).
\end{aligned}
\end{equation} 

The ROC AUC measures the area under the Receiver Operating Characteristic (ROC) curve, which plots the True Positive Rate (TPR) against the False Positive Rate (FPR) at various decision thresholds. The PR AUC represents the area under the Precision–Recall (PR) curve, depicting the trade-off between Precision and Recall across thresholds. Both metrics provide threshold-independent evaluations of classifier performance, particularly useful for imbalanced datasets.

\subsection{Implementation Details}

{
All experiments were implemented in Python~3.8 using PyTorch~1.13.1 on a Linux~5.4.0 . 
The proposed model was trained on a workstation equipped with two NVIDIA GeForce RTX~5090D GPUs (24~GB~$\times$~2). 
All experiments were conducted with a fixed random seed of 0 to ensure reproducibility.
All evaluation metrics are computed by aggregating predictions over the entire test set. We report the \textbf{mean ± standard} deviation to reflect the variability of model performance across individual samples.

\paragraph{\textbf{Training Configuration}}
All models were trained for 200 epochs with a batch size of 32 and L2 regularization ($\lambda = 1\times10^{-3}$). 
The AdamW optimizer was employed with an initial learning rate of $1\times10^{-4}$ and a weight decay of $1\times10^{-5}$. 
A Cosine Annealing Learning Rate (LR) scheduler was adopted to gradually reduce the learning rate during training. 
The loss function was Cross-Entropy Loss, and all models were trained end-to-end in mixed precision on GPUs.
\textbf{Training strategy}: Stage 1 (Independent Pre-training):
Each backbone network (ConvNeXt-Base, ResNet34, and ResNet18) is trained independently. All parameters are trainable in this stage.Stage 2 (Fusion Training with Frozen Backbones):
After loading the pretrained weights, we construct the fusion model and freeze all backbone parameters. Only the parameters of the fusion classification head are optimized. 

\paragraph{\textbf{Data Preprocess}} 
All CT volumes were reconstructed from the original DICOM data with automatic rescale correction to ensure standardized Hounsfield Unit (HU) representation. No additional intensity windowing, clipping, histogram matching, or scanner-specific normalization was applied, thereby preserving the native HU distribution. The original voxel spacing of each scan was retained without resampling. For each clustered nodule, a fixed-size 3D cube (64 × 64 × 56 voxels) was extracted around the averaged annotation centroid with boundary control. No explicit lung segmentation or region masking was performed, and each patch includes both the lesion and its peri-nodular context.

\paragraph{\textbf{Data augmentation}} was performed
to improve robustness, consisting of random vertical flips, random affine transformations (0--180\textdegree\ rotation), random perspective distortions, and thin plate spline transformations, each applied with a probability of 0.5. 
No additional scanner-specific harmonization or intensity re-scaling was applied beyond native HU reconstruction.
\paragraph{\textbf{Fair Comparison Protocol}} 
All baseline models were trained and evaluated using the identical dataset split and the same nodule-centered 3D patches described in Data Preprocess, without model-specific preprocessing or additional normalization. Official implementations were adopted whenever available, and only necessary input-dimension adjustments were made. All models were trained under comparable optimization settings, including the same optimizer type, learning rate search range, training epochs, and data augmentation strategy, with hyperparameters selected based on validation performance. For 2D-based architectures, 3D patches were decomposed into axial slices and processed slice-wise, with slice-level features aggregated to obtain a case-level representation.

\paragraph{\textbf{Few-shot fine-tuning for multi-center generalization}}
This strategy supports generalization across data centers while preserving previously learned representations.
To reflect the limited data availability and center-specific variations in the target setting, we adopt a few-shot fine-tuning.
We employed 20 positive–negative sample pairs from a public LIDC–IDRI dataset as a replay mechanism to alleviate catastrophic forgetting, together with 20 positive–negative pairs from a private USTC-FHLN dataset for few-shot fine-tuning. This joint strategy enables effective adaptation to the target domain while preserving previously learned representations.
The two datasets originate from independent institutions with no patient overlap, and that all fine-tuning samples are strictly disjoint from the corresponding test sets. 
}


\begin{table}[t]
\centering
\caption{Quantitative results of benign and malignant classification under different crop sizes.}
\label{Table1}
renewcommand{\arraystretch}{1.3}
\resizebox{0.47\textwidth}{!}{
\begin{tabular}{ c c c c c c } 
\Xhline{1.2pt}
\cellcolor{CadetBlue!20} Crop Size &\cellcolor{CadetBlue!20} Model &\cellcolor{CadetBlue!20} Accuracy(\%)\( \uparrow \) &\cellcolor{CadetBlue!20} Precision(\%)\( \uparrow \) &\cellcolor{CadetBlue!20} Recall\( \uparrow \) &\cellcolor{CadetBlue!20} F1-score\( \uparrow \)  \\
\Xhline{1.2pt}
\multirow{5}{*}{\shortstack{[32,32,56]}}
& \cellcolor{gray!10}ConvNeXt-T \cite{liu2022convnet} & \cellcolor{gray!10}82.96 & \cellcolor{gray!10}82.08 & \cellcolor{gray!10}82.96 & \cellcolor{gray!10}82.27 \\
& ConvNeXt-B \cite{liu2022convnet}   & 85.78 & 85.23 & 85.78 & 85.34 \\
&\cellcolor{gray!10} ConvNeXt-XL \cite{liu2022convnet}  &\cellcolor{gray!10} 83.26 &\cellcolor{gray!10} 82.33 &\cellcolor{gray!10} 83.26 &\cellcolor{gray!10} 82.43 \\
& ResNet-18 \cite{he2016deep}  & 82.67 & 81.67 & 82.67 & 81.81 \\
& \cellcolor{gray!10}ResNet-34 \cite{he2016deep}  & \cellcolor{gray!10}82.22 & \cellcolor{gray!10}81.25 & \cellcolor{gray!10}82.22 & \cellcolor{gray!10}81.47 \\
\Xhline{1.2pt}
\multirow{5}{*}{\shortstack{[48,48,56]}}
& ConvNeXt-T \cite{liu2022convnet} & 81.63 & 80.30 & 81.63 & 80.13 \\
& \cellcolor{gray!10}ConvNeXt-B \cite{liu2022convnet}   & \cellcolor{gray!10}83.26 & \cellcolor{gray!10}82.48 & \cellcolor{gray!10}83.26 & \cellcolor{gray!10}82.67 \\
& ConvNeXt-XL \cite{liu2022convnet}  & 83.56 & 83.00 & 83.56 & 83.20 \\
& \cellcolor{gray!10}ResNet-18 \cite{he2016deep}  & \cellcolor{gray!10}82.37 & \cellcolor{gray!10}81.76 & \cellcolor{gray!10}82.37 & \cellcolor{gray!10}81.99 \\
& ResNet-34 \cite{he2016deep}  & 81.78 & 80.61 & 81.78 & 80.76 \\
\Xhline{1.2pt} 
\multirow{5}{*}{[64,64,56]} 
& \cellcolor{gray!10}ConvNeXt-T \cite{liu2022convnet} & \cellcolor{gray!10}81.19 & \cellcolor{gray!10}79.92 & \cellcolor{gray!10}81.19 & \cellcolor{gray!10}78.76 \\
& ConvNeXt-B \cite{liu2022convnet}   & 83.85 & 83.01 & 83.85 & 82.57 \\
&\cellcolor{gray!10} ConvNeXt-XL \cite{liu2022convnet}  &\cellcolor{gray!10} 81.19 &\cellcolor{gray!10} 80.46 &\cellcolor{gray!10} 81.19 &\cellcolor{gray!10} 80.73 \\
& ResNet-18 \cite{he2016deep}  & 82.52 & 81.58 & 82.52 & 81.78 \\
& \cellcolor{gray!10}ResNet-34 \cite{he2016deep}  & \cellcolor{gray!10}83.26 & \cellcolor{gray!10}82.35 & \cellcolor{gray!10}83.26 & \cellcolor{gray!10}82.48 \\
\Xhline{1.2pt}
\multirow{5}{*}{\shortstack{[80,80,56]}}
& ConvNeXt-T \cite{liu2022convnet} & 81.04 & 79.66 & 81.04 & 78.73 \\
& \cellcolor{gray!10}ConvNeXt-B \cite{liu2022convnet}   & \cellcolor{gray!10}81.78 & \cellcolor{gray!10}80.87 & \cellcolor{gray!10}81.78 & \cellcolor{gray!10}81.14 \\
& ConvNeXt-XL \cite{liu2022convnet}  & 80.15 & 78.52 & 80.15 & 77.36 \\
& \cellcolor{gray!10}ResNet-18 \cite{he2016deep}  & \cellcolor{gray!10}81.19 & \cellcolor{gray!10}80.46 & \cellcolor{gray!10}81.19 & \cellcolor{gray!10}80.73 \\
& ResNet-34 \cite{he2016deep}  & 83.11 & 82.56 & 83.11 & 82.76 \\
\Xhline{1.2pt}
\multirow{5}{*}{[96,96,56]} 
& \cellcolor{gray!10}ConvNeXt-T \cite{liu2022convnet} & \cellcolor{gray!10}81.93 & \cellcolor{gray!10}80.66 & \cellcolor{gray!10}81.93 & \cellcolor{gray!10}80.39 \\
& ConvNeXt-B \cite{liu2022convnet}   & 83.26 & 82.31 & 83.26 & 82.38 \\
&\cellcolor{gray!10} ConvNeXt-XL \cite{liu2022convnet}  &\cellcolor{gray!10} 81.48 &\cellcolor{gray!10} 80.71 &\cellcolor{gray!10} 81.48 &\cellcolor{gray!10} 80.98 \\
& ResNet-18 \cite{he2016deep}  & 82.96 & 82.54 & 82.96 & 82.72 \\
& \cellcolor{gray!10}ResNet-34 \cite{he2016deep}  & \cellcolor{gray!10}82.37 & \cellcolor{gray!10}81.61 & \cellcolor{gray!10}82.37 & \cellcolor{gray!10}81.85 \\
\Xhline{1.2pt}
\end{tabular}}
\end{table}

\section{Experimental Results}\label{section:Experimental Results}
\subsection{Main Result}

\subsubsection{The impact of crop size on classification results}

As listed in Table~\ref{Table1}, different crop sizes produce significant effects on the classification performance of benign and malignant nodules.
\paragraph{\textbf{Obs. \ding{182}: 32D: Core fine grained structure}}
Models trained with a receptive field of [32, 32, 56] achieve competitive and sometimes superior performance compared with models that use larger receptive fields. For example, the accuracy of ConvNeXt-B reaches 85.78\%. This finding highlights the importance of high resolution information such as edge clarity, lobulation, internal heterogeneity and subtle density variations.
\paragraph{\textbf{Obs. \ding{183}: 64D: Most balanced organizational scale}}
Most models reach a stable peak or near peak in overall performance with the [64, 64, 56] configuration. For instance, ConvNeXt-B achieves 83.85\% and ResNet-34 reaches 83.26\%. This indicates that a moderately enlarged field of view enables the model to capture peri nodule information such as vascular convergence, pleural adjacency and boundary distortion while avoiding excessive background noise.
\paragraph{\textbf{Obs. \ding{184}: 96D: Wide area context and spatial relationships}}
Model performance remains stable at this scale and ResNet-18 even reaches its highest accuracy of 82.96\%. This suggests that semantic level cues derived from long range anatomical context also contribute to malignancy assessment. Features such as large scale vascular connectivity, tissue compression patterns and global spatial arrangements become more prominent at this scale and complement the information captured at smaller fields of view.

In summary, we select the three input sizes of [32, 32, 56], [64, 64, 56] and [96, 96, 56] to form a \textbf{progressive hierarchical structure} that transitions from fine structural details to local semantics and then to global spatial relationships. This design enables the model to use complementary information across different spatial levels for benign and malignant classification.
\subsubsection{Comparative experiment on lung nodule classification}

\paragraph{\textbf{Obs. \ding{182}: Optimal Performance in Quantitative Metrics}} 
As shown in Table \ref{Table2}, the proposed method achieves consistently superior performance on both the LIDC-IDRI and USTC-FHLN datasets. On LIDC-IDRI, it attains the best results across all nine metrics, with an accuracy of \textbf{86.96\%}, surpassing ResNet-18 (82.67\%) and DeepLung (83.70\%). The precision and recall reach \textbf{86.56\%} and \textbf{86.96\%}, yielding the highest F1-score of \textbf{86.66\%}, while the balanced accuracy of \textbf{80.12\%} exceeds that of MVCS (79.52\%), indicating stable class-wise performance. In addition, the ROC-AUC (\textbf{85.40\%}) and PR-AUC (\textbf{87.27\%}) outperform transformer-based methods such as ViT-B and the DINO series. On the USTC-FHLN dataset, the proposed method again achieves the highest accuracy (\textbf{84.24\%}) and F1-score (\textbf{84.25\%}), together with a high precision of \textbf{86.07\%} and balanced accuracy of \textbf{85.19\%}. Although slightly lower ROC-AUC or PR-AUC values are observed compared to some methods, the proposed approach maintains a better balance between accuracy and robustness, demonstrating strong generalization across different datasets.


\begin{figure*}[t]
\centering
\includegraphics[width=0.9\textwidth]{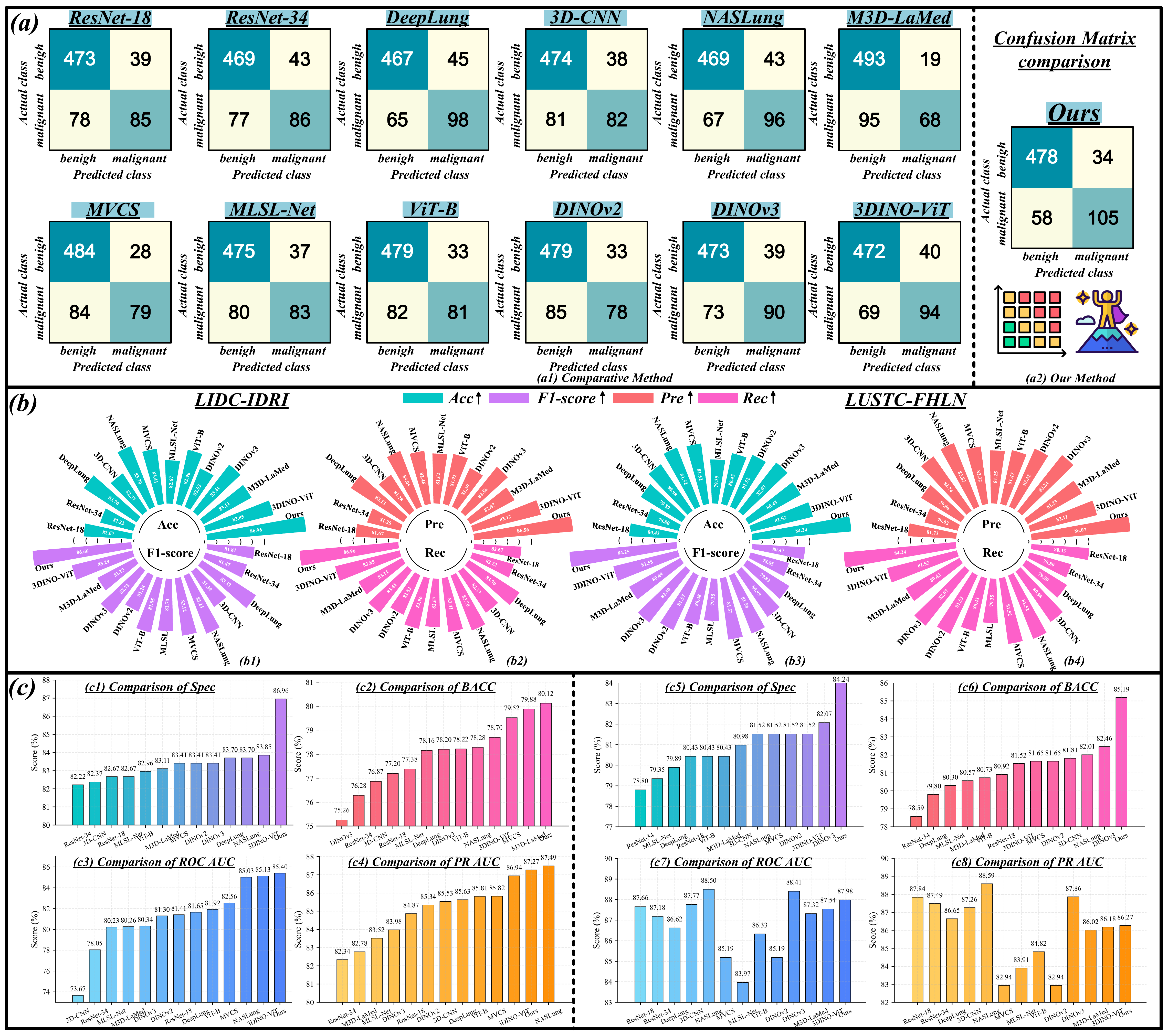}
\caption{(a) Confusion matrices for pulmonary nodule classification on LIDC-IDRI dataset; 
(b) Accuracy, F1-score, Precision, and Recall comparison on LIDC-IDRI and USTC-FHLN datasets; 
(c) Specificity, Balanced Accuracy, ROC AUC, and PR AUC visualization on LIDC-IDRI and USTC-FHLN datasets. 
Performance comparison of various methods for benign-malignant pulmonary nodule classification.}
\label{ThreePhaseCompare_V12.0.pdf}
\end{figure*}

\begin{figure*}[!t]
\centering
\includegraphics[width=0.96\textwidth]{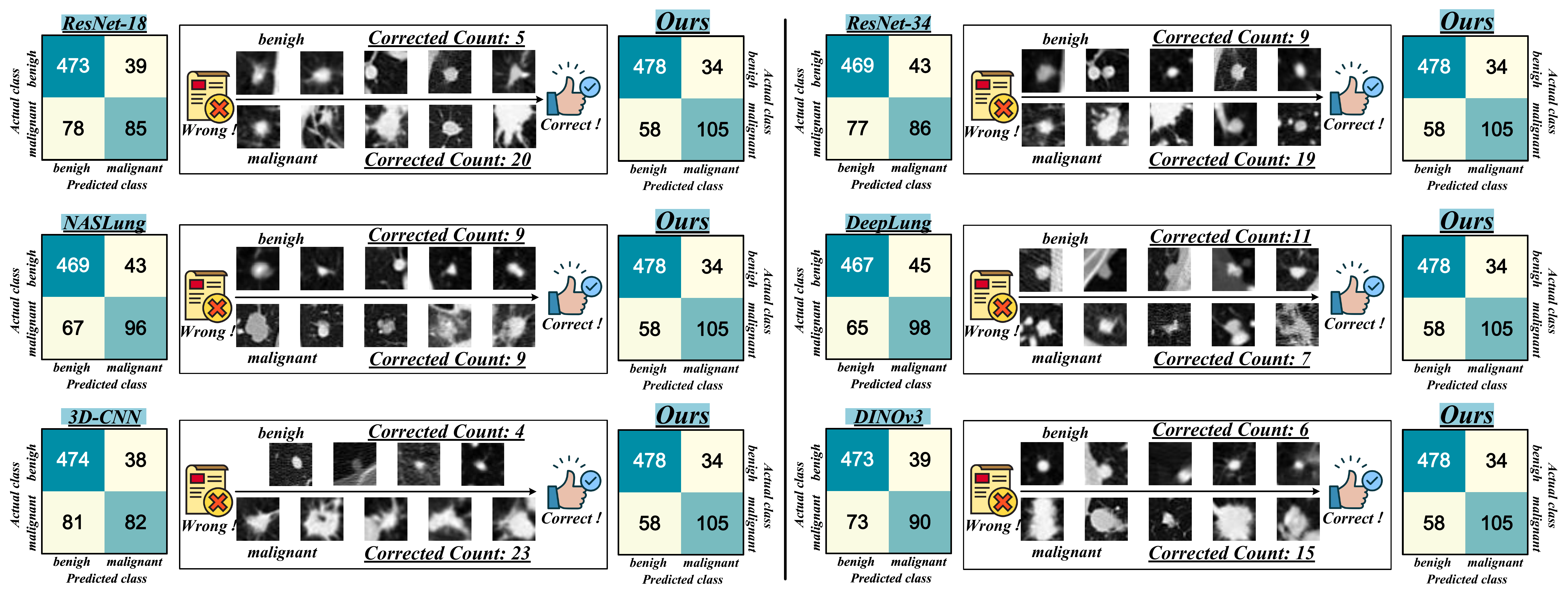}
\caption{The comparison results illustrate that our method consistently outperforms competing methods by effectively correcting a substantial portion of misclassified and incorrectly predicted samples. In particular, the proposed approach demonstrates a stronger ability to recover correct predictions in challenging cases.}
\label{compareV3}
\end{figure*}

\begin{table*}[t]
\centering
\caption{Quantitative comparison results on the LIDC-IDRI and USTC-FHLN dataset. The bold values indicate the best performance.}
\label{Table2}
\resizebox{0.87\textwidth}{!}{
{ \renewcommand{\arraystretch}{1.3}
\begin{tabular}{ c c c c c c c c c c} 
\Xhline{1.2pt}
\rowcolor{CadetBlue!20}
Dataset & Model & Acc(\%)\( \uparrow \) & Pre(\%)\( \uparrow \) & Rec(\%)\( \uparrow \) & Spec(\%)\( \uparrow \) & F1(\%)\( \uparrow \) & BACC(\%)\( \uparrow \) & ROC(\%)\( \uparrow \) & PR(\%)\( \uparrow \) \\
\Xhline{1.2pt}
\multirow{11}{*}{\shortstack{LIDC-\\IDRI}}
& \cellcolor{gray!10}ResNet-18 \cite{he2016deep}  & \cellcolor{gray!10}82.67 $\pm$ 1.58 & \cellcolor{gray!10}81.67 $\pm$ 4.16 & \cellcolor{gray!10}82.67 $\pm$ 3.24 & \cellcolor{gray!10}82.67 & \cellcolor{gray!10}81.81 $\pm$ 3.52 & \cellcolor{gray!10}77.20 & \cellcolor{gray!10}81.41 & 84.87 \\
& ResNet-34 \cite{he2016deep}  & 82.22 $\pm$ 1.50 & 81.25 $\pm$ 4.26 & 82.22 $\pm$ 3.17 & 82.22 & 81.47 $\pm$ 3.53 & 76.28 & 78.05 & 82.34 \\
& \cellcolor{gray!10}DeepLung \cite{zhu2018deeplung}  & \cellcolor{gray!10}83.70 $\pm$ 1.41 & \cellcolor{gray!10}83.13 $\pm$ 3.90 & \cellcolor{gray!10}83.70 $\pm$ 3.80 & \cellcolor{gray!10}83.70 & \cellcolor{gray!10}83.33 $\pm$ 3.16 & \cellcolor{gray!10}78.16 & \cellcolor{gray!10}81.65 & \cellcolor{gray!10}85.63 \\
& 3D-CNN \cite{zunair2020uniformizing}  & 82.37 $\pm$ 1.45 & 81.28 $\pm$ 4.27 & 82.37 $\pm$ 3.13 & 82.37 & 81.39 $\pm$ 3.48 & 76.87 & 73.67 & 85.53 \\
&\cellcolor{gray!10} NASLung \cite{jiang2021learning}  &\cellcolor{gray!10} 83.70 $\pm$ 1.42 &\cellcolor{gray!10} 83.05 $\pm$ 3.94 &\cellcolor{gray!10} 83.70 $\pm$ 3.82 &\cellcolor{gray!10} 83.70 &\cellcolor{gray!10} 83.24 $\pm$ 3.19 &\cellcolor{gray!10} 78.28 &\cellcolor{gray!10} 85.03 &\cellcolor{gray!10} \textbf{87.49} \\
& MVCS \cite{zhu2022multi} & 83.41 $\pm$ 1.50 & 82.46 $\pm$ 3.91 & 83.41 $\pm$ 3.90 & 83.41 & 82.12 $\pm$ 3.35 & 79.52 & 82.56 & 85.82 \\
& \cellcolor{gray!10}MLSL-Net \cite{yi2022multi} & \cellcolor{gray!10}82.67 $\pm$ 1.57 & \cellcolor{gray!10}81.62 $\pm$ 4.18 & \cellcolor{gray!10}82.67 $\pm$ 3.27 & \cellcolor{gray!10}82.67 & \cellcolor{gray!10}81.70 $\pm$ 3.50 & \cellcolor{gray!10}77.38 & \cellcolor{gray!10}80.23 & \cellcolor{gray!10}83.52 \\
& ViT-B \cite{wu2020visual} & 82.96 $\pm$ 1.55 & 81.92 $\pm$ 4.13 & 82.96 $\pm$ 3.18 & 82.96 & 81.85 $\pm$ 3.45 & 78.22 & 81.92 & 85.81 \\
& \cellcolor{gray!10}DINOv2 \cite{oquab2023dinov2} & \cellcolor{gray!10}82.52 $\pm$ 1.52 & \cellcolor{gray!10}81.39 $\pm$ 4.23 & \cellcolor{gray!10}82.52 $\pm$ 3.28 & \cellcolor{gray!10}82.52 & \cellcolor{gray!10}81.28 $\pm$ 3.54 & \cellcolor{gray!10}77.60 & \cellcolor{gray!10}78.33 & \cellcolor{gray!10}82.88 \\
& DINOv3 \cite{simeoni2025dinov3} & 83.41 $\pm$ 1.49 & 82.56 $\pm$ 3.88 & 83.41 $\pm$ 1.50 & 83.41 & 82.71 $\pm$ 3.41 & 78.20 & 81.30 & 85.34 \\
& \cellcolor{gray!10}M3D-LaMed \cite{bai2024m3d} & \cellcolor{gray!10}83.11 $\pm$  1.46 & \cellcolor{gray!10}82.47 $\pm$  4.05 & \cellcolor{gray!10}83.11 $\pm$ 3.82 & \cellcolor{gray!10}83.11 & \cellcolor{gray!10}81.13 $\pm$ 3.51 & \cellcolor{gray!10}79.88 & \cellcolor{gray!10}80.26 & \cellcolor{gray!10}82.78 \\
&  3DINO-ViT\cite{xu2025generalizable} & 83.85 $\pm$ 1.39 & 83.12 $\pm$ 3.90 & 83.85 $\pm$ 3.78 & 83.85 & 83.29 $\pm$ 3.20 & 78.70 & 85.13 & 86.94 \\
\Xcline{2-10}{1.2pt}
\Xcline{2-10}{1.2pt}
\rowcolor{yellow!20} \cellcolor{white} &   Ours    & \textbf{86.96} $\pm$ 1.28 & \textbf{86.56} $\pm$ 3.73 & \textbf{86.96} $\pm$ 3.88 & \textbf{86.96} & \textbf{86.66} $\pm$ 3.12 & \textbf{80.12} & \textbf{85.40} & 87.27 \\
\Xhline{1.2pt}
\multirow{11}{*}{\shortstack{USTC-\\FHLN}}
& \cellcolor{gray!10}ResNet-18 \cite{he2016deep}  & \cellcolor{gray!10}80.43 $\pm$ 3.38 & \cellcolor{gray!10}81.73 $\pm$ 3.16 & \cellcolor{gray!10}80.43 $\pm$ 4.70 & \cellcolor{gray!10}80.43 & \cellcolor{gray!10}80.47 $\pm$ 3.28 & \cellcolor{gray!10}80.92 & \cellcolor{gray!10}87.66 & \cellcolor{gray!10}87.84 \\
& ResNet-34 \cite{he2016deep}  & 78.80 $\pm$ 3.59
 & 79.02 $\pm$ 3.62
 & 78.80 $\pm$ 4.88
 & 78.80 & 78.85 $\pm$ 3.44
 & 78.59 & 87.18 & 87.49 \\
& \cellcolor{gray!10}DeepLung \cite{zhu2018deeplung}  & \cellcolor{gray!10}79.89 $\pm$ 3.42
 & \cellcolor{gray!10}79.86 $\pm$ 3.39
 & \cellcolor{gray!10}79.89 $\pm$ 4.80 & \cellcolor{gray!10}79.89 & \cellcolor{gray!10}79.82 $\pm$ 3.33 & \cellcolor{gray!10}79.80 & \cellcolor{gray!10}86.62 & \cellcolor{gray!10}86.65 \\
& 3D-CNN \cite{zunair2020uniformizing}  & 80.98 $\pm$ 3.30 & 82.74 $\pm$ 3.05 & 80.98 $\pm$ 4.64 & 80.98 & 80.99 $\pm$ 3.20 & 81.81 & 87.77 & 87.26 \\
&\cellcolor{gray!10} NASLung \cite{jiang2021learning}  &\cellcolor{gray!10} 81.52 $\pm$ 3.23 &\cellcolor{gray!10} 82.83 $\pm$ 3.00 &\cellcolor{gray!10} 81.52 $\pm$ 4.60 &\cellcolor{gray!10} 81.52 &\cellcolor{gray!10} 81.56 $\pm$ 3.17 &\cellcolor{gray!10} 82.01 &\cellcolor{gray!10} \textbf{88.50} &\cellcolor{gray!10} \textbf{88.59} \\
& MVCS \cite{zhu2022multi} & 81.52 $\pm$ 3.22 & 82.32 $\pm$ 3.06 & 81.52 $\pm$ 4.59 & 81.52 & 81.57 $\pm$ 3.14 & 81.65 & 85.19 & 82.94 \\
& \cellcolor{gray!10}MLSL-Net \cite{yi2022multi} & \cellcolor{gray!10}79.35 $\pm$ 3.48 & \cellcolor{gray!10}81.25 $\pm$ 3.25 & \cellcolor{gray!10}79.35 $\pm$ 4.81 & \cellcolor{gray!10}79.35 & \cellcolor{gray!10}79.35 $\pm$ 3.36 & \cellcolor{gray!10}80.30 & \cellcolor{gray!10}83.97 & \cellcolor{gray!10}83.91 \\
& ViT-B \cite{wu2020visual} & 80.43 $\pm$ 3.39 & 81.47 $\pm$ 3.21 & 80.43 $\pm$ 4.71 & 80.43 & 80.48 $\pm$ 3.27 & 80.73 & 86.33 & 84.82 \\
& \cellcolor{gray!10}DINOv2 \cite{oquab2023dinov2} & \cellcolor{gray!10}81.52 $\pm$ 3.24 & \cellcolor{gray!10}82.32 $\pm$ 3.10 & \cellcolor{gray!10}81.52 $\pm$ 4.59 & \cellcolor{gray!10}81.52 & \cellcolor{gray!10}81.57 $\pm$ 3.16 & \cellcolor{gray!10}81.65 & \cellcolor{gray!10}85.19 & \cellcolor{gray!10}82.94 \\
& DINOv3 \cite{simeoni2025dinov3} & 82.07 $\pm$ 3.08 & 83.24 $\pm$ 2.83 & 82.07 $\pm$ 4.28 & 82.07 & 82.10 $\pm$ 2.93 & 82.46 & 88.41 & 87.86 \\
& \cellcolor{gray!10}M3D-LaMed \cite{bai2024m3d} & \cellcolor{gray!10}80.43 $\pm$ 3.39 & \cellcolor{gray!10}81.23 $\pm$ 3.25 & \cellcolor{gray!10}80.43 $\pm$ 4.72 & \cellcolor{gray!10}80.43 & \cellcolor{gray!10}80.49 $\pm$ 3.28 & \cellcolor{gray!10}80.57 & \cellcolor{gray!10}87.32 & \cellcolor{gray!10}86.02 \\
&  3DINO-ViT\cite{xu2025generalizable} & 81.52 $\pm$ 3.23 & 82.11 $\pm$ 3.18 & 81.52 $\pm$ 4.59 & 81.52 & 81.58 $\pm$ 3.15 & 81.52 & 87.54 & 86.18 \\
\Xcline{2-10}{1.2pt}
\Xcline{2-10}{1.2pt}
\rowcolor{yellow!20} \cellcolor{white} & Ours & \textbf{84.24} $\pm$ 2.71 & \textbf{86.07} $\pm$ 2.59 & \textbf{84.24} $\pm$ 4.24 & \textbf{84.24} & \textbf{84.25} $\pm$ 2.87 & \textbf{85.19} & 87.98 & 86.27 \\
\Xhline{1.2pt}
\end{tabular}}
}
\end{table*}
\paragraph{\textbf{Obs. \ding{183}: Clinical Relevance of Model Behavior}} Beyond numerical improvements, the model behavior aligns with clinical relevance. Table two shows that different receptive field sizes emphasize complementary diagnostic cues. 
\textbf{\ding{171} Macro-scale inputs} introduce broader anatomical context such as tissue deformation and long range vascular connections, which further inform malignancy interpretation.
\textbf{\ding{163} Meso-scale inputs} provide a balanced view by capturing peri nodule patterns including vascular convergence and pleural relationships while maintaining sufficient structural detail. 
\textbf{\ding{168} Micro-scale inputs} focus on high resolution nodule characteristics such as margin sharpness, lobulation, internal texture variations and subtle density changes that radiologists commonly rely on for malignancy risk estimation. 
To summarize, the proposed method benefits from a \textbf{hierarchical structure design}, enabling the model to leverage clinical features across \textbf{multiple spatial levels}.


\paragraph{\textbf{$\blacksquare$ Visualization Analysis}}
The visualization results in Figure~\ref{ThreePhaseCompare_V12.0.pdf} illustrate the superiority of our method across multiple evaluation perspectives. \textbf{\ding{182} As shown in Figure~\ref{ThreePhaseCompare_V12.0.pdf} (a)}, the confusion matrices show that our model achieves fewer misclassifications in both benign and malignant categories compared with CNN-based methods (ResNet-18/34, 3D-CNN) and Transformer-based approaches (ViT-B, DINOv2/3). This indicates stronger discriminative capability, especially for challenging borderline nodules.
\textbf{\ding{183} As shown in Figure~\ref{ThreePhaseCompare_V12.0.pdf} (b)}, our method consistently outperforms all compared models in Accuracy, Precision, Recall, and F1-score. Specifically, it achieves the highest Acc (86.96\%) and F1 (86.66\%), demonstrating effective reduction in both false positives and false negatives while maintaining stable classification performance. 
\textbf{\ding{184} As shown in Figure~\ref{ThreePhaseCompare_V12.0.pdf} (c)}, the results further confirm the robustness of our approach. Our method achieves the best Specificity, Balanced Accuracy, ROC AUC (85.40\%), and PR AUC (87.27\%), indicating superior generalization and more stable performance under varying decision thresholds.
Consistent with the quantitative results in Table~\ref{Table2}, our approach delivers leading performance across all major metrics, validating its effectiveness for benign–malignant pulmonary nodule classification on the LIDC-IDRI and USTC-FHLN dataset.

Figure \ref{compareV3} provides a qualitative comparison between our method and several representative baselines. As illustrated, conventional models such as ResNet-18, ResNet-34, and 3D-CNN tend to misclassify challenging cases, particularly confusing benign nodules with malignant ones when local appearance cues are ambiguous. In contrast, our method successfully corrects these errors in multiple examples. For instance, several nodules that are incorrectly predicted as malignant by ResNet-based models are correctly identified as benign by our approach, while false benign predictions made by DeepLung and NASLung are also effectively rectified. These examples demonstrate that our method leverages richer contextual information to resolve ambiguities that cannot be addressed by local features alone, resulting in more reliable and robust lung nodule classification.

\begin{table}[htbp]
\centering
\caption{Network module ablation experiment and results analysis on LIDC-IDRI Dataset. (The bold indicators indicate the best performance)}
{ \renewcommand{\arraystretch}{1.3}
\resizebox{0.47\textwidth}{!}{
\begin{tabular}{ c c c c c c c c }
\Xhline{1.2pt}
\multirow{2}{*}{Version} & \multicolumn{3}{c}{\cellcolor{CadetBlue!20} Experiments} & \multicolumn{4}{c}{\cellcolor{CadetBlue!20}M$^3$Net} \\
&\cellcolor{CadetBlue!20} LSF &\cellcolor{CadetBlue!20} HSF &\cellcolor{CadetBlue!20} Transformer 
&\cellcolor{CadetBlue!20} Acc(\%)\( \uparrow \) 
&\cellcolor{CadetBlue!20} Pre(\%)\( \uparrow \) 
&\cellcolor{CadetBlue!20} Rec(\%)\( \uparrow \) 
&\cellcolor{CadetBlue!20} F1(\%)\( \uparrow \) \\
\Xhline{1.2pt}

a & \checkmark & & &
\cellcolor{gray!10} $85.78_{\color{red}{\downarrow 1.18}}$ &
\cellcolor{gray!10} $85.17_{\color{red}{\downarrow 1.39}}$ &
\cellcolor{gray!10} $85.78_{\color{red}{\downarrow 1.18}}$ &
\cellcolor{gray!10} $84.96_{\color{red}{\downarrow 1.70}}$ \\

b & & \checkmark & &
$85.48_{\color{red}{\downarrow 1.48}}$ &
$84.85_{\color{red}{\downarrow 1.71}}$ &
$85.48_{\color{red}{\downarrow 1.48}}$ &
$84.60_{\color{red}{\downarrow 2.06}}$ \\

c & & & \checkmark &
\cellcolor{gray!10} $86.37_{\color{red}{\downarrow 0.59}}$ &
\cellcolor{gray!10} $85.94_{\color{red}{\downarrow 0.62}}$ &
\cellcolor{gray!10} $86.37_{\color{red}{\downarrow 0.59}}$ &
\cellcolor{gray!10} $86.06_{\color{red}{\downarrow 0.60}}$ \\

d & \checkmark & \checkmark & &
$86.67_{\color{red}{\downarrow 0.29}}$ &
$86.15_{\color{red}{\downarrow 0.41}}$ &
$86.67_{\color{red}{\downarrow 0.29}}$ &
$86.10_{\color{red}{\downarrow 0.56}}$ \\

e & \checkmark & & \checkmark &
\cellcolor{gray!10} $85.93_{\color{red}{\downarrow 1.03}}$ &
\cellcolor{gray!10} $85.42_{\color{red}{\downarrow 1.14}}$ &
\cellcolor{gray!10} $85.93_{\color{red}{\downarrow 1.03}}$ &
\cellcolor{gray!10} $84.95_{\color{red}{\downarrow 1.71}}$ \\

f & & \checkmark & \checkmark &
$86.52_{\color{red}{\downarrow 0.44}}$ &
$86.02_{\color{red}{\downarrow 0.54}}$ &
$86.52_{\color{red}{\downarrow 0.44}}$ &
$86.08_{\color{red}{\downarrow 0.58}}$ \\

g & \checkmark & \checkmark & \checkmark &
\cellcolor{yellow!20}\textbf{86.96} &
\cellcolor{yellow!20}\textbf{86.56} &
\cellcolor{yellow!20}\textbf{86.96} &
\cellcolor{yellow!20}\textbf{86.66} \\

\Xhline{1.2pt}
\end{tabular}}
}
\label{Table3}
\end{table}

\subsection{Module Ablation Experiment.}

{LSF (Low-Scale Features) corresponds to the feature representation $\mathbf{T}_1$ in Eq. (\ref{eq:transformer_block1}), capturing fine-grained, local information. HSF (High-Scale Features) corresponds to the feature representation $\mathbf{T}_2$ in Eq. (\ref{eq:transformer_block2}), encoding more global, coarse-grained information.}
Table \ref{Table3} presents the ablation results of M$^3$Net under different module combinations. \ding{168} Using only LSF (version a) or HSF (version b) module yields comparable performance, indicating that both low-scale and high-scale features contribute to representation learning. The separate introduction of Transformer (version c) further improves the accuracy to 86.37\%, highlighting its ability to capture long-range dependencies. \ding{163} When LSF and HSF are combined (version d), the accuracy rises to 86.67\%, confirming the complementarity of the interaction between high and low-scale features. Pairwise fusion with Transformer (versions e and f) brings additional performance improvements, demonstrating effective feature fusion. \ding{171} Notably, integrating all three components (LSF + HSF + Transformer, version g) achieves the best performance with an accuracy of \textbf{86.96\%}, a precision of \textbf{86.56\%}, a recall of \textbf{86.96\%} and an F1-score of \textbf{86.66\%}. This suggests that the joint use of \textbf{multi-scale features} and \textbf{Transformer} modeling enables M$^3$Net to capture richer and more discriminative representations.

\subsection{Qualitative Analysis and Discussion}

\paragraph{\textbf{Obs. \ding{182}: Performance Comparison of Single-Scale Inputs}}
The ablation results in Table \ref{Table4} clearly demonstrate the performance discrepancy among different single-scale inputs. Version a (only $\bm{X}_{32}$) achieves an accuracy (Acc) of 85.78\%, outperforming Version b (only $\bm{X}_{64}$, 82.96\%) and Version c (only $\bm{X}_{96}$, 84.00\%). This validates our core motivation that \ding{168} \textbf{32$\times$32$\times$56} receptive field captures high-resolution details critical for clinical assessment, edge sharpness, lobulation, internal heterogeneity and subtle density variations of nodules are fully retained, providing direct visual evidence for malignancy judgment. In contrast, Version b. \ding{163} \textbf{64$\times$64$\times$56} with a moderate field of view, leverages information around the nodules (such as vascular convergence, pleural proximity) but is affected by a certain degree of background noise, leading to a 2.82\% Acc drop compared to Version a. Although Version c. \ding{171} \textbf{96$\times$96$\times$56} captures long-range anatomical clues like large-scale vascular connectivity and tissue compression patterns, the lack of fine details still results in a 1.78\% lower Acc than Version a.
\paragraph{\textbf{Obs. \ding{183}: Synergistic Effect of Multi-Scale Inputs}}
Multi-scale combinations consistently outperform single-scale counterparts, with Version g (integrating $\bm{X}_{32}$, $\bm{X}_{64}$ and $\bm{X}_{96}$) achieving the optimal performance (Acc: \textbf{86.96\%}, Precision (Pre): \textbf{86.56\%}, Recall (Rec): \textbf{86.96\%}, F1: \textbf{86.66\%}). Specifically, Version g surpasses Version a (single $\bm{X}_{32}$) by 1.18\% in Acc and 1.32\% in F1, confirming the complementary value of multi-scale information. Versions d ($\bm{X}_{32}+\bm{X}_{64}$) and e ($\bm{X}_{32}+\bm{X}_{96}$) yield Acc of 83.85\% and 85.33\% respectively, both lower than Version g. This indicates that the progressive hierarchical structure, which ranges from fine structure ($\bm{X}_{32}$) to local semantics ($\bm{X}_{64}$, capturing vascular convergence and pleural proximity) and global spatial relationships ($\bm{X}_{96}$, reflecting overall anatomical arrangement), is indispensable for comprehensive malignancy evaluation.
\begin{table}[h]
\centering
\caption{Ablation study results of M$^3$Net of 3P input on LIDC-IDRI Dataset. (The bold indicators indicate the best performance)}
{ \renewcommand{\arraystretch}{1.3}
\resizebox{0.48\textwidth}{!}{
\begin{tabular}{ c c c c c c c c }
\Xhline{1.2pt}
\multirow{2}{*}{Version} 
& \multicolumn{3}{c}{\cellcolor{CadetBlue!20} Experiments} 
& \multicolumn{4}{c}{\cellcolor{CadetBlue!20} M$^3$Net} \\
&\cellcolor{CadetBlue!20} $\bm{X}_{32}$ 
&\cellcolor{CadetBlue!20} $\bm{X}_{64}$ 
&\cellcolor{CadetBlue!20} $\bm{X}_{96}$ 
&\cellcolor{CadetBlue!20} Acc(\%)\( \uparrow \) 
&\cellcolor{CadetBlue!20} Pre(\%)\( \uparrow \) 
&\cellcolor{CadetBlue!20} Rec(\%)\( \uparrow \) 
&\cellcolor{CadetBlue!20} F1\( \uparrow \) \\
\Xhline{1.2pt}

a & \checkmark & & &
\cellcolor{gray!10} $85.78_{\color{red}{\downarrow 1.18}}$ &
\cellcolor{gray!10} $85.23_{\color{red}{\downarrow 1.33}}$ &
\cellcolor{gray!10} $85.78_{\color{red}{\downarrow 1.18}}$ &
\cellcolor{gray!10} $85.34_{\color{red}{\downarrow 1.32}}$ \\

b & & \checkmark & &
$82.96_{\color{red}{\downarrow 4.00}}$ &
$82.12_{\color{red}{\downarrow 4.44}}$ &
$82.96_{\color{red}{\downarrow 4.00}}$ &
$82.32_{\color{red}{\downarrow 4.34}}$ \\

c & & & \checkmark &
\cellcolor{gray!10} $84.00_{\color{red}{\downarrow 2.96}}$ &
\cellcolor{gray!10} $83.44_{\color{red}{\downarrow 3.12}}$ &
\cellcolor{gray!10} $84.00_{\color{red}{\downarrow 2.96}}$ &
\cellcolor{gray!10} $83.63_{\color{red}{\downarrow 3.03}}$ \\

d & \checkmark & \checkmark & &
$83.85_{\color{red}{\downarrow 3.11}}$ &
$83.84_{\color{red}{\downarrow 2.72}}$ &
$83.85_{\color{red}{\downarrow 3.11}}$ &
$81.70_{\color{red}{\downarrow 4.96}}$ \\

e & \checkmark & & \checkmark &
\cellcolor{gray!10} $85.33_{\color{red}{\downarrow 1.63}}$ &
\cellcolor{gray!10} $84.86_{\color{red}{\downarrow 1.70}}$ &
\cellcolor{gray!10} $85.33_{\color{red}{\downarrow 1.63}}$ &
\cellcolor{gray!10} $84.11_{\color{red}{\downarrow 2.55}}$ \\

f & & \checkmark & \checkmark &
$82.67_{\color{red}{\downarrow 4.29}}$ &
$82.56_{\color{red}{\downarrow 4.00}}$ &
$82.67_{\color{red}{\downarrow 4.29}}$ &
$80.02_{\color{red}{\downarrow 6.64}}$ \\

g & \checkmark & \checkmark & \checkmark &
\cellcolor{yellow!20}\textbf{86.96} &
\cellcolor{yellow!20}\textbf{86.56} &
\cellcolor{yellow!20}\textbf{86.96} &
\cellcolor{yellow!20}\textbf{86.66} \\

\Xhline{1.2pt}
\end{tabular}}
}
\label{Table4}
\end{table}

\paragraph{\textbf{Obs. \ding{184}: Clinical Significance Interpretation}}
To sum up, the design of this \textbf{progressive hierarchical structure} successfully simulates the logic of clinical diagnosis. \ding{168} It starts with the analysis of global anatomical relationships, \ding{163} proceeds to the evaluation of local contextual signs such as vascular attachment and pleural indentation, \ding{171} and finally incorporates the assessment of fine internal structures of nodules including calcification distribution and spiculation. This design enables the model to mimic the clinical diagnostic thinking process and thus improve the accuracy of diagnosis.
{
\subsection{Ablation Study on Clinically-Constrained Data Augmentation}

\begin{table}[h]
\centering
\caption{Ablation study of clinically-constrained data augmentation on M$^3$Net (3P input) over LIDC-IDRI.}
{ \renewcommand{\arraystretch}{1.3}
\resizebox{0.48\textwidth}{!}{
\begin{tabular}{ c c c c c }
\Xhline{1.2pt}
\multirow{2}{*}{Configuration}  
& \multicolumn{4}{c}{\cellcolor{CadetBlue!20} Performance (\%)} \\
&\cellcolor{CadetBlue!20} Acc(\%)\( \uparrow \) 
&\cellcolor{CadetBlue!20} Pre(\%)\( \uparrow \) 
&\cellcolor{CadetBlue!20} Rec(\%)\( \uparrow \) 
&\cellcolor{CadetBlue!20} F1\( \uparrow \) \\
\Xhline{1.2pt}

Without Augmentation &
\cellcolor{gray!10} $86.37_{\color{red}{\downarrow 0.59}}$ &
\cellcolor{gray!10} $85.89_{\color{red}{\downarrow 0.67}}$ &
\cellcolor{gray!10} $86.37_{\color{red}{\downarrow 0.59}}$ &
\cellcolor{gray!10} $85.98_{\color{red}{\downarrow 0.68}}$ \\

With Augmentation (Ours)&
\cellcolor{yellow!20}\textbf{86.96} &
\cellcolor{yellow!20}\textbf{86.56} &
\cellcolor{yellow!20}\textbf{86.96} &
\cellcolor{yellow!20}\textbf{86.66} \\

\Xhline{1.2pt}
\end{tabular}}
}
\label{Table-ablation}
\end{table}

Table \ref{Table-ablation} presents the ablation results of the proposed clinically-constrained data augmentation. Without augmentation, M$^3$Net achieves 86.37\% accuracy and an F1-score of 85.98\%. Applying the augmentation consistently improves all metrics, reaching 86.96\% accuracy and 86.66\% F1-score. The modest yet consistent gains ($\approx$0.6–0.7\%) indicate improved generalization rather than shortcut learning or distribution-shift artifacts, supporting the clinical plausibility of the augmentation strategy.
}

\subsection{Grad-CAM–based Qualitative Attention Visualization}

\begin{figure*}[!t]
\centering
\includegraphics[width=0.96\textwidth]{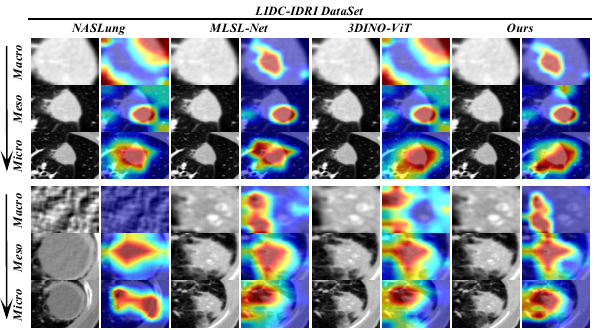}
\caption{Grad-CAM visualizations for two representative cases. For each case, the original CT slice and the corresponding attention maps generated by different models are shown. Warmer colors indicate regions contributing more to the model prediction. The visualizations highlight how the models focus on the nodule regions and surrounding contextual structures during classification. 
Displayed slices correspond to each model’s peak attention.
}
\label{CAM}
\end{figure*}

{
As illustrated in Fig. \ref{CAM}, Grad-CAM visualizations are presented for three representative cases across different models and input scales. The baseline methods NASLung and MLSL-Net show relatively scattered activation patterns, where the highlighted regions are sometimes distributed over surrounding lung tissues rather than being consistently concentrated on the nodule area. The 3DINO-ViT model demonstrates improved localization in some cases; however, its attention responses remain relatively diffuse and may extend beyond the lesion boundaries. In contrast, the proposed M3Net tends to produce more stable and spatially concentrated responses around the nodule regions across different scales. This behavior suggests that the multi-scale macro–meso–micro representation and hierarchical fusion mechanism may help the model capture both local texture cues and broader contextual information during prediction. These visualizations provide qualitative insight into how different models attend to nodule-related structures during classification.
}

\subsection{Efficiency Analysis}

\begin{table}[h]
\centering
\caption{Efficiency Comparison of M$^3$Net and Baseline Models (GFLOPs and Parameters). The bold indicates the best performance}
\label{table5} 
\resizebox{0.48\textwidth}{!}{
{ \renewcommand{\arraystretch}{1.3}
\begin{tabular}{ c c c c}
\Xhline{1.2pt}
\cellcolor{CadetBlue!20}Model & \cellcolor{CadetBlue!20}Gflops ($\times 10^9$) & \cellcolor{CadetBlue!20}Parameters (M) & \cellcolor{CadetBlue!20}Accuracy(\%)\( \uparrow \) \\
\Xhline{1.2pt}
\cellcolor{gray!10}NASLung \cite{jiang2021learning} & \cellcolor{gray!10}21.335 & \cellcolor{gray!10}28.456 & \cellcolor{gray!10}83.70$_{\color{red}{\downarrow 3.26}}$ \\
DeepLung \cite{zhu2018deeplung} & 45.316 & 36.968 & 83.70$_{\color{red}{\downarrow 3.26}}$ \\
\cellcolor{gray!10}MVCS \cite{zhu2022multi} & \cellcolor{gray!10}48.202 & \cellcolor{gray!10}98.987 & \cellcolor{gray!10}83.41$_{\color{red}{\downarrow 3.55}}$ \\
DinoV3 \cite{simeoni2025dinov3} & 539.150 & 85.904 & 83.41$_{\color{red}{\downarrow 3.55}}$ \\
\Xhline{1.2pt}
\cellcolor{yellow!20}Ours & \cellcolor{yellow!20}\textbf{20.079} & \cellcolor{yellow!20}184.608 & \cellcolor{yellow!20}\textbf{86.96} \\
\Xhline{1.2pt}
\end{tabular}}
}
\end{table}

{As shown in Table \ref{table5}, the proposed M$^3$Net achieves the best classification accuracy of 86.96\%, outperforming all baseline models by a clear margin. Compared with NASLung, DeepLung, MVCS, and DinoV3, our method improves accuracy by 3.26\%–3.55\%, demonstrating superior predictive capability.
In terms of computational complexity, M$^3$Net requires 20.079 GFLOPs, which is lower than DeepLung, MVCS, and DinoV3, and even slightly lower than NASLung (21.335 GFLOPs). This indicates that our model maintains competitive computational efficiency while achieving substantially better performance. 
While M$^3$Net involves a relatively larger number of parameters, this is unlikely to pose practical challenges for medical deployment, where diagnostic accuracy is of greater importance. Nonetheless, further improving parameter efficiency could be considered in future work to enhance overall model compactness.
}
\subsection{Ablation Study on Loss Function Components}

\begin{table}[h]
\centering
\caption{Ablation Study on the Four Loss Components of M$^3$Net. (The bold indicators indicate the best performance)}
{ \renewcommand{\arraystretch}{1.3}
\resizebox{0.48\textwidth}{!}{
\begin{tabular}{ c c c c c c c c c }
\Xhline{1.2pt}
\multirow{2}{*}{Version} 
& \multicolumn{4}{c}{\cellcolor{CadetBlue!20}Loss Components} 
& \multicolumn{4}{c}{\cellcolor{CadetBlue!20} M$^3$Net} \\
& \cellcolor{CadetBlue!20}$\mathcal{L}_{\text{align}}$ 
& \cellcolor{CadetBlue!20}$\mathcal{L}_{\text{cov}}$ 
& \cellcolor{CadetBlue!20}$\mathcal{L}_{\text{orth}}$ 
& \cellcolor{CadetBlue!20}$\mathcal{L}_{\text{nuc}}$
& \cellcolor{CadetBlue!20}Acc(\%)\( \uparrow \) 
& \cellcolor{CadetBlue!20}Pre(\%)\( \uparrow \) 
& \cellcolor{CadetBlue!20}Rec(\%)\( \uparrow \) 
& \cellcolor{CadetBlue!20}F1\( \uparrow \) \\
\Xhline{1.2pt}

a & \checkmark & & & &
\cellcolor{gray!10}84.89$_{\color{red}{\downarrow 2.07}}$ &
\cellcolor{gray!10}84.21$_{\color{red}{\downarrow 2.35}}$ &
\cellcolor{gray!10}84.89$_{\color{red}{\downarrow 2.07}}$ &
\cellcolor{gray!10}83.82$_{\color{red}{\downarrow 2.84}}$ \\

b & & \checkmark & & &
84.59$_{\color{red}{\downarrow 2.37}}$ &
83.96$_{\color{red}{\downarrow 2.60}}$ &
84.59$_{\color{red}{\downarrow 2.37}}$ &
84.12$_{\color{red}{\downarrow 2.54}}$ \\

c & & & \checkmark & &
\cellcolor{gray!10}84.00$_{\color{red}{\downarrow 2.96}}$ &
\cellcolor{gray!10}83.16$_{\color{red}{\downarrow 3.40}}$ &
\cellcolor{gray!10}84.00$_{\color{red}{\downarrow 2.96}}$ &
\cellcolor{gray!10}82.81$_{\color{red}{\downarrow 3.85}}$ \\

d &  &  &  & \checkmark &
80.59$_{\color{red}{\downarrow 6.37}}$ &
80.03$_{\color{red}{\downarrow 6.53}}$ &
80.59$_{\color{red}{\downarrow 6.37}}$ &
80.26$_{\color{red}{\downarrow 6.40}}$ \\



e & \checkmark & \checkmark & \checkmark & \checkmark &
\cellcolor{yellow!20}\textbf{86.96} &
\cellcolor{yellow!20}\textbf{86.56} &
\cellcolor{yellow!20}\textbf{86.96} &
\cellcolor{yellow!20}\textbf{86.66} \\

\Xhline{1.2pt}
\end{tabular}}
}
\label{Table4}
\end{table}

{
Table~\ref{Table4} presents an ablation study on the four loss components of M$^3$Net. Using only the full alignment loss $\mathcal{L}{\text{align}}$ (Version a) drops F1 by 2.84\% compared to the complete model (Version g). Individually, the covariance loss $\mathcal{L}{\text{cov}}$ (b) and orthogonality loss $\mathcal{L}{\text{orth}}$ (c) yield moderate declines in F1 by 2.54\% and 3.85\%, respectively. The nuclear norm loss $\mathcal{L}{\text{nuc}}$ (d) contributes most significantly, with a 6.40\% F1 drop when used alone. The combination of all four losses (e) achieves the highest performance across all metrics, demonstrating that each component complements the others, and the nuclear norm is particularly critical for maintaining global consistency in cross-scale representation alignment.
}


\section{Conclusions}
\label{section:conclusion}
{
We propose M$^{3}$Net, a novel Macro$\rightarrow$Meso$\rightarrow$Micro hierarchical 3D network for pulmonary nodule classification that 
inspired
by mirroring radiologists' top-down diagnostic reasoning. Extensive experiments on both the public LIDC-IDRI dataset and clinical USTC-FHLN dataset, M$^{3}$Net achieves state-of-the-art performance,
surpassing all baseline methods. 
By progressively integrating global anatomical context, peri-nodular regional information, and fine-grained structural details through cross-scale feature alignment, the proposed framework provides a structured representation of multi-scale information that is consistent with clinical observation patterns. While this design is motivated by clinical reasoning and may facilitate model interpretability, further dedicated studies are needed to quantitatively evaluate its explainability. Overall, M$^{3}$Net offers an effective and clinically motivated framework for pulmonary nodule classification.
\textbf{Generalization.} Although this study focuses on pulmonary nodule classification, the proposed macro–meso–micro hierarchical reasoning framework is not limited to this specific task. In principle, it can be extended to other lesion-centric medical imaging problems, such as liver lesion analysis and breast mass assessment, where multi-scale contextual cues and structured diagnostic reasoning are equally important. The alignment between model inference and clinically meaningful hierarchical context may provide a generalizable paradigm for interpretable lesion analysis.
\textbf{Limitation.} Nevertheless, the applicability of the framework to other clinical scenarios may be influenced by several factors, including data heterogeneity across imaging modalities and institutions, variations in annotation protocols, and differences in lesion appearance and scale. These factors may affect the stability of hierarchical feature representations and require task-specific adaptation or additional training data to ensure robust generalization. In future work, we plan to conduct reader studies with radiologists to systematically evaluate the clinical relevance and reliability of the model explanations.
}

\section{Acknowledgments}\label{sec:acknowledge}
This research was supported by the Single Disease Lung Cancer Research Innovation Project of Hefei Cancer Hospital, Chinese Academy of Sciences (Y24ZL0010101).


\bibliographystyle{cas-model2-names}

\bibliography{cas-refs}



\end{document}